\documentclass{article}

\usepackage{amsmath,amsfonts,bm}

\def\eqref#1{equation~\ref{#1}}
\def\1{\bm{1}}

\def\rva{{\mathbf{a}}}

\def\rvg{{\mathbf{g}}}

\def\rvs{{\mathbf{s}}}

\def\mM{{\bm{M}}}

\DeclareMathAlphabet{\mathsfit}{\encodingdefault}{\sfdefault}{m}{sl}
\SetMathAlphabet{\mathsfit}{bold}{\encodingdefault}{\sfdefault}{bx}{n}
\newcommand{\tens}[1]{\bm{\mathsfit{#1}}}

\def\tG{{\tens{G}}}

\def\gA{{\mathcal{A}}}

\def\gD{{\mathcal{D}}}

\def\gL{{\mathcal{L}}}

\def\gP{{\mathcal{P}}}

\def\gR{{\mathcal{R}}}
\def\gS{{\mathcal{S}}}
\def\gT{{\mathcal{T}}}

\def\sM{{\mathbb{M}}}

\def\sT{{\mathbb{T}}}

\newcommand{\E}{\mathbb{E}}

\newcommand{\R}{\mathbb{R}}

\newcommand{\transformer}{\mathrm{Transformer}}

\DeclareMathOperator*{\argmax}{arg\,max}
\DeclareMathOperator*{\argmin}{arg\,min}

\newcommand{\te}[1]{\texttt{#1}}

\usepackage{microtype}
\usepackage{graphicx}
\usepackage{subfigure}
\usepackage{booktabs} % for professional tables

\usepackage{hyperref}

\usepackage{enumitem}

\usepackage[accepted]{icml2024}

\usepackage{amsmath}
\usepackage{amssymb}
\usepackage{mathtools}
\usepackage{amsthm}
\usepackage{color}
\usepackage{xcolor}
\definecolor{grey}{rgb}{0.5,0.5,0.5}
\usepackage[capitalize,noabbrev]{cleveref}

\usepackage{tabularray} 
\usepackage{multirow}
\usepackage{multicol}
\usepackage{rotating} % for rotating text (vertical text)
\usepackage{array}
\usepackage[hang]{footmisc}
\usepackage{pifont}
\usepackage{duckuments}
\usepackage{enumitem}

\usepackage[makeroom]{cancel}

\theoremstyle{plain}
\newtheorem{theorem}{Theorem}[section]

\theoremstyle{definition}
\newtheorem{definition}[theorem]{Definition}

\theoremstyle{remark}

\usepackage[textsize=tiny]{todonotes}

\newcommand{\PreserveBackslash}[1]{\let\temp=\\#1\let\\=\temp}
\newcolumntype{C}[1]{>{\PreserveBackslash\centering}p{#1}}
\newcolumntype{R}[1]{>{\PreserveBackslash\raggedleft}p{#1}}
\newcolumntype{L}[1]{>{\PreserveBackslash\raggedright}p{#1}}

\icmltitlerunning{HarmoDT: Harmony Multi-Task Decision Transformer for Offline Reinforcement Learning}

\begin{document}

\twocolumn[
\icmltitle{HarmoDT: Harmony Multi-Task Decision Transformer for\\ Offline Reinforcement Learning}



\icmlsetsymbol{equal}{*}

\begin{icmlauthorlist}
\icmlauthor{Shengchao Hu}{equal,sjtu,pjlab}
\icmlauthor{Ziqing Fan}{equal,sjtu,pjlab}
\icmlauthor{Li Shen}{zs,jd}
\icmlauthor{Ya Zhang}{sjtu,pjlab}
\icmlauthor{Yanfeng Wang}{sjtu,pjlab}
\icmlauthor{Dacheng Tao}{ntu}
\end{icmlauthorlist}

\icmlaffiliation{sjtu}{Shanghai Jiao Tong University, China}
\icmlaffiliation{pjlab}{Shanghai AI Laboratory, China}
\icmlaffiliation{zs}{Sun Yat-sen University, China}
\icmlaffiliation{jd}{JD Explore Adademy, China}
\icmlaffiliation{ntu}{Nanyang Technological University, Singapore}

\icmlcorrespondingauthor{Li Shen}{mathshenli@gmail.com}

\icmlkeywords{Machine Learning, ICML}

\vskip 0.3in
]



\printAffiliationsAndNotice{\icmlEqualContribution} % otherwise use the standard text.

\begin{abstract}
    The purpose of offline multi-task reinforcement learning (MTRL) is to develop a unified policy applicable to diverse tasks without the need for online environmental interaction.
    Recent advancements approach this through sequence modeling, leveraging the Transformer architecture's scalability and the benefits of parameter sharing to exploit task similarities.
    However, variations in task content and complexity pose significant challenges in policy formulation, necessitating judicious parameter sharing and management of conflicting gradients for optimal policy performance.
    In this work, we introduce the Harmony Multi-Task Decision Transformer (HarmoDT), a novel solution designed to identify an optimal harmony subspace of parameters for each task. 
    We approach this as a bi-level optimization problem, employing a meta-learning framework that leverages gradient-based techniques.
    The upper level of this framework is dedicated to learning a task-specific mask that delineates the harmony subspace, while the inner level focuses on updating parameters to enhance the overall performance of the unified policy.
    Empirical evaluations on a series of benchmarks demonstrate the superiority of HarmoDT, verifying the effectiveness of our approach.
\end{abstract}

\begin{figure}[ht!]
\includegraphics[width=0.46\textwidth]{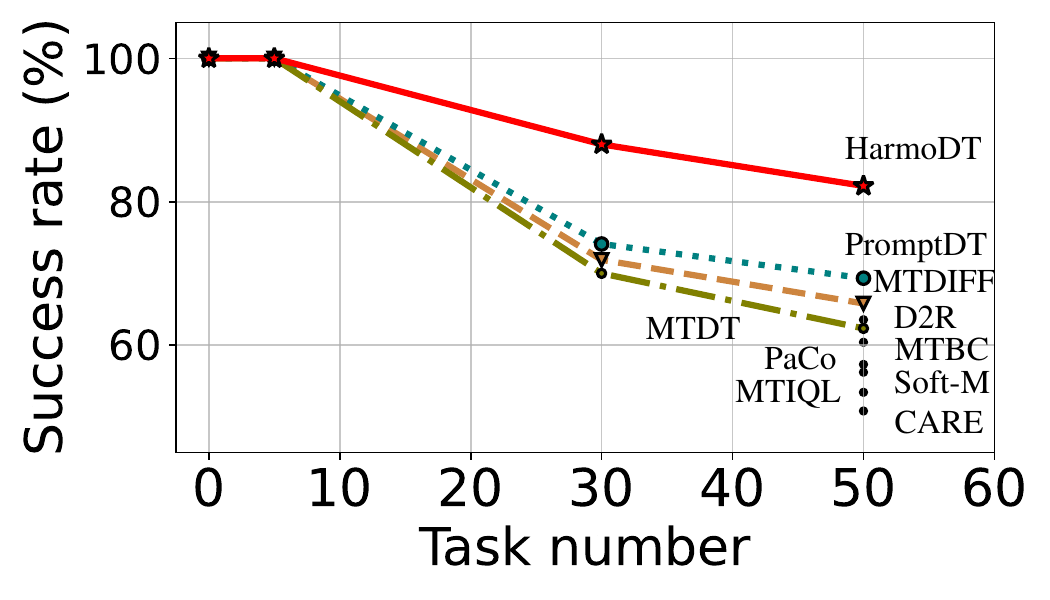}
\vspace{-0.2cm}
\caption{
Illustration of a comparative analysis of success rates across various task numbers within the Meta-World benchmark, focusing on prevalent MTRL algorithms. 
An in-depth exploration of these results refers to Section~\ref{sec:exp}.
}
\label{fig:performance_num}
\vspace{-0.2cm}
\end{figure}
\section{Introduction}

Offline reinforcement learning (RL) \citep{levine2020offline} enables the learning of policies directly from an existing offline dataset, thus eliminating the need for interaction with the actual environment.
Despite the promising developments of offline RL in various robotic tasks, its successes have been largely confined to individual tasks within specific domains, such as locomotion or manipulation \citep{fu2020d4rl, CQL}.
Drawing inspiration from human learning capabilities, where individuals often acquire new skills by building upon existing ones and spend less time mastering similar tasks, there's a growing interest in the potential of training a set of tasks with inherent similarities in a more cohesive and efficient manner \citep{lee2022multi}.
This perspective leads to the exploration of multi-task reinforcement learning (MTRL), which seeks to develop a versatile policy to address a variety of tasks.

Recent developments in Offline RL, such as the Decision Transformer \citep{DT} and Trajectory Transformer \citep{TT}, have abstracted offline RL as a sequence modeling (SM) problem, showcasing their ability to transform extensive datasets into powerful decision-making tools \citep{TRL}. 
These models are particularly beneficial for multi-task RL challenges, offering a high-capacity framework capable of accommodating task variances and assimilating extensive knowledge from diverse datasets. 
Additionally, they open up possibilities for integrating advancements \citep{brown2020language} from language modeling into MTRL methodologies.
However, the application of these high-capacity sequential models to MTRL presents considerable algorithmic challenges.
As indicated by \citet{yu2020meta}, simply employing a shared network backbone for all diverse robot manipulation tasks can lead to severe gradient conflicts.
This situation arises when the gradient direction for a particular task starkly contrasts with the majority consensus direction.
Such unregulated sharing of parameters and their optimization under conflicting gradient conditions can contravene the foundational goals of MTRL, degrading performance relative to task-specific training methods \citep{sun2022paco}.
Furthermore, the issue of gradient conflict is exacerbated by an increase in the number of tasks (detailed in Section \ref{sec:rethink}), underscoring the urgency for effective solutions to these challenges.

Existing works on offline MTRL generally address the problem in one of three ways \citep{sun2022paco}: 
1) developing shared structures for the sub-policies of different tasks, as explored in works by \citet{calandriello2014sparse, yang2020multi, lin2022switch}; 
2) optimizing task-specific representations to condition the policies, as discussed by \citet{sodhani2021multi, lee2022multi, he2023diffusion};
3) addressing the conflicting gradients arising from different task losses during training, a focus of research by \citet{yu2020gradient, chen2020just, liu2021conflict}.
While these methods have demonstrated effectiveness in different scenarios, they often fall short of adequately addressing the occurrence of conflicting gradients that stem from indiscriminate parameter sharing \citep{guangyuan2022recon}.
In contrast, our innovative method, the Harmony Multi-Task Decision Transformer (HarmoDT), diverges from these traditional approaches. 
HarmoDT endeavors to identify a harmony parameter subspace within a single policy for each task, offering a novel solution to the challenges of offline MTRL.

\begin{figure}[t!]
\centering 
\subfigure[Conflicting during MTRL.]{
\label{fig:intro_4}
\includegraphics[width=0.245\textwidth]{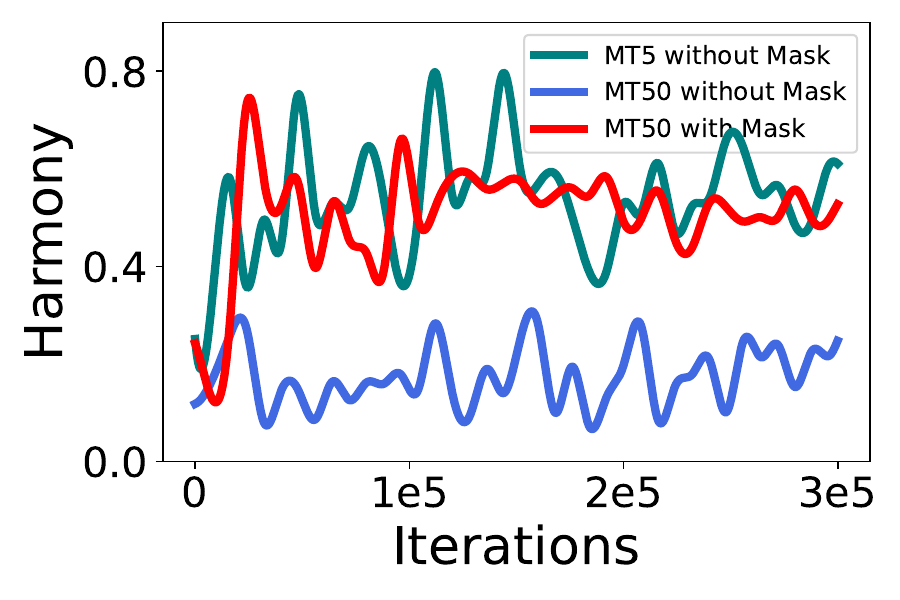}}\!\!\!
\subfigure[Success with task masks.]{
\label{fig:intro_5}
\includegraphics[width=0.21\textwidth]{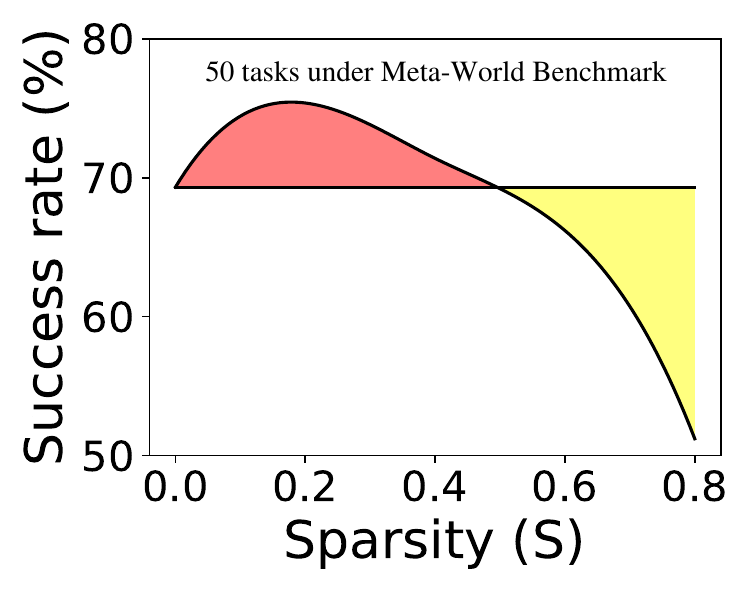}}
\vspace{-0.2cm}
\caption{Illustration of the harmony degree among trainable weights during training for policies with and without randomly initialized masks (left panel), and the success rates achieved when applying masks with varying sparsity levels (right panel).}
\label{fig:redun_conflict}
\vspace{-0.2cm}
\end{figure}

To reduce the occurrence of the conflicting gradient, the idea of adopting distinct parameter subspaces for each task is straightforward.
Empirical observations, depicted by Figure~\ref{fig:intro_4}, affirm that the application of masks significantly mitigates conflicts, leading to considerable performance gains across various sparsity ratios\footnote{Sparsity ratio refers to the percentage of inactive weights.}, as contrasted with the non-mask baseline shown in Figure~\ref{fig:intro_5}.
Building upon these insights, our HarmoDT seeks to identify an optimal harmony subspace of parameters for each task by incorporating trainable task-specific masks during MTRL training. 
This approach is conceptualized as a bi-level optimization problem, employing a meta-learning framework to discern the harmony subspace mask via gradient-based techniques.
At the upper level, we focus on learning a task-specific mask that delineates the harmony subspace, while at the inner level, we update parameters to augment the collective performance of the unified model under the guidance of the task-specific mask.
Empirical evaluations of HarmoDT, conducted across a broad spectrum of tasks in both seen and unseen settings, demonstrate its efficacy against multiple state-of-the-art algorithms.
Additionally, we provide extensive ablation studies on various aspects, including scalability, model size, hyper-parameters, and visualizations, to comprehensively validate our approach.

In summary, our research makes three significant contributions to the field of MTRL:
\begin{itemize}[leftmargin=15pt]
    \item We rethink the challenges in MTRL from the perspective of sequence modeling, analyze gradient conflicts with increasing task numbers, and propose the harmony subspace using task-specific masks (Section \ref{sec:rethink}).
    \item We model the problem as a bi-level optimization problem and introduce a meta-learning framework to find the optimal harmony subspace mask through gradient-based techniques~(Section \ref{sec:method}).
    \item We demonstrate the superior performance of HarmoDT through rigorous testing on a broad spectrum of benchmarks, establishing its state-of-the-art effectiveness in MTRL scenarios (Section \ref{sec:exp}).
\end{itemize}
\section{Preliminary}

\subsection{Offline Reinforcement Learning}

The goal of RL is to learn a policy $\pi_{\theta}(\rva | \rvs)$ maximizing the expected cumulative discounted rewards $\E[\sum_{t=0}^{\infty} \gamma^t \gR(\rvs_t, \rva_t)]$ in a Markov decision process (MDP), which is a six-tuple $(\gS, \gA, \gP, \gR, \gamma, d_0)$, with state space $\gS$, action space $\gA$, environment dynamics $\gP(\rvs' | \rvs, \rva): \gS \times \gS \times \gA \rightarrow [0,1]$, reward function $\gR: \gS \times \gA \rightarrow \R$, discount factor $\gamma \in [0, 1)$, and initial state distribution $d_0$ \citep{sutton2018reinforcement}.
The action-value or Q-value of a policy $\pi$ is defined as $Q^{\pi}(\rvs_t, \rva_t) = \E_{\rva_{t+1}, \rva_{t+2} \dots \sim \pi} [\sum_{i=0}^{\infty} \gamma^i \gR(\rvs_{t+i}, \rva_{t+i})]$.
In the offline setting \citep{levine2020offline}, instead of the online environment, a static dataset $\gD = \{(\rvs, \rva, \rvs', r)\}$, collected by a behavior policy $\pi_{\beta}$, is provided.
Offline RL algorithms learn a policy entirely from this static offline dataset $\gD$, without any online interactions with the environment.

In the multi-task setting, different tasks can have different reward functions, state spaces, and transition functions. 
We consider all tasks to share the same action space with the same embodied agent.
Given a specific task $\gT \sim p(\gT)$, a task-specified MDP can be defined as $(\gS^{\gT}, \gA^{\gT}, \gP^{\gT}, \gR^{\gT}, \gamma, d_0^{\gT})$.
Instead of solving a single MDP, the goal of multi-task RL is to find an optimal policy that maximizes expected return over all the tasks: $\pi^* = \argmax_{\pi} \E_{\gT \sim p(\gT)} \E_{\rva_t \sim \pi} [\sum_{t=0}^{\infty} \gamma^t r_t^{\gT}] $.
The static dataset $\gD$ correspondingly is partitioned into per-task subsets as $\gD = \cup_{i=1}^N \gD_i$, where $N$ is the number of tasks.

\subsection{Prompt Decision Transformer}
The integration of the Transformer \citep{transformer} architecture in offline RL for SM has gained prominence in recent years. 
Studies in NLP reveal that Transformers pre-trained on extensive datasets exhibit notable few-shot or zero-shot learning capabilities within a prompt-based framework \citep{liu2023pre, brown2020language}.
Building on this, Prompt-DT adapts the prompt-based methodology to offline RL, facilitating few-shot generalization to novel tasks.
Unlike NLP, where prompts are typically text-based and adapt to various tasks through blank-filling formats, Prompt-DT introduces trajectory prompts. 
These prompts consist of state, action, and return-to-go tuples $(\rvs^*, \rva^*, \hat{r}^*)$, providing directed guidance to RL agents with few-shot demonstrations.
Each element marked with the superscript $\cdot^*$ is relevant to the trajectory prompt.
Note that the length of the trajectory prompt is usually shorter than the task's horizon, encompassing only essential information to facilitate task identification, yet inadequate for complete task imitation.
During training with offline collected data, Prompt-DT utilizes $\tau_{i,t}^{input}=(\tau_i^*, \tau_{i,t})$ as input for each task $\gT_i$. 
Here, $\tau^{input}_{i,t}$ consists of the $K^*$-step trajectory prompt $\tau_i^*$ and the most recent $K$-step history $\tau_{i,t}$, and is formulated as:
\begin{align}
\small
\label{eq:input}
    &\tau^{input}_{i, t} = (\hat{r}^*_{i, 1}, \rvs^*_{i, 1}, \rva^*_{i, 1}, \dots,  \hat{r}^*_{i, K^*}, \rvs^*_{i, K^*}, \rva^*_{i, K^*}, \nonumber \\ 
    &\quad\hat{r}_{i, t-K+1}, \rvs_{i, t-K+1}, \rva_{i, t-K+1}, \dots,  \hat{r}_{i, t}, \rvs_{i, t}, \rva_{i, t}).
\end{align}
The prediction head linked to a state token $\rvs$ is designed to predict the corresponding action $\rva$.
For continuous action spaces, the training objective aims to minimize the mean-squared loss:
\begin{equation}
\small
   \gL_{DT}\! =\! \mathbb{E}_{\tau^{input}_{i, t} \sim \gD_i} \left[ \frac{1}{K}\!\! \sum_{m=t-K+1}^t\!\! (\rva_{i, m} - \pi(\tau^*_i, \tau_{i,m} ) )^2 \right]. 
\end{equation}

\begin{figure*}[t!]
\centering
\includegraphics[width=0.98\textwidth]{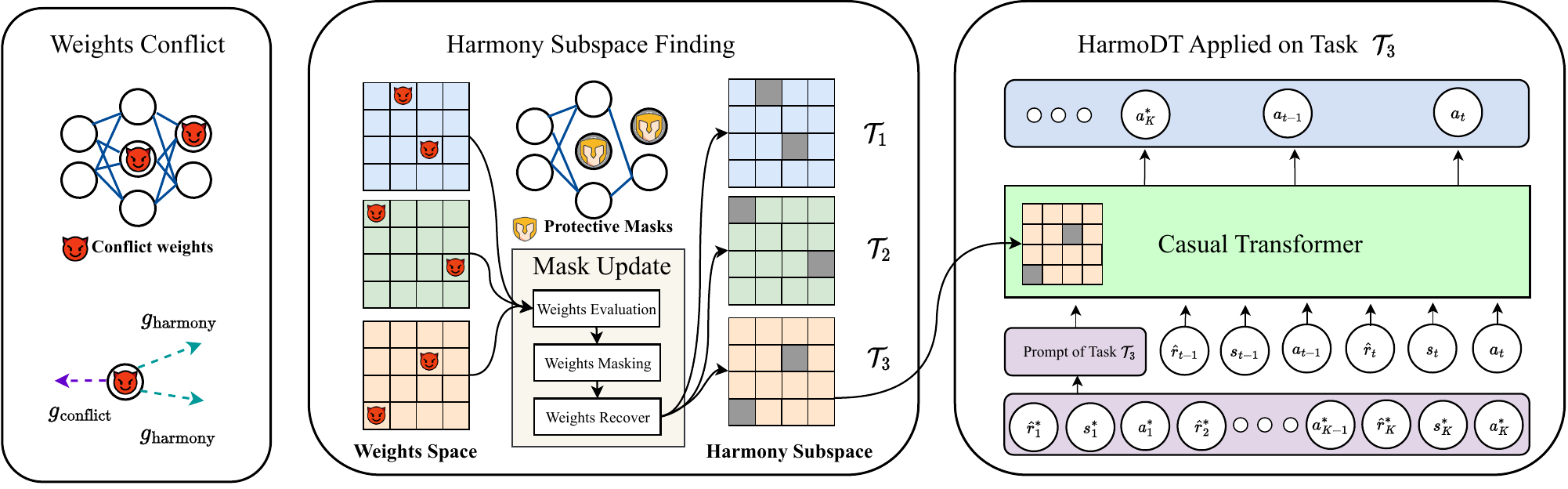}
\vspace{-0.2cm}
\caption{Illustration of the conflicting problem and the framework of our method HarmoDT to find a harmony subspace for each task. 
The left panel shows the conflicting phenomenon reflected by divergent task-specific gradients.
The middle panel illustrates the procedure to find a harmony subspace for each task via the strategic learning of task masks.
The right panel demonstrates the workflow of HarmoDT based on the DT architecture with prompts and learned harmony subspace of weights when handling a task, such as $\gT_3$.}
\label{fig:pipeline}
\vspace{-0.3cm}
\end{figure*}

\section{Rethinking SM with MTRL}
\label{sec:rethink}

Recent works in offline RL conceptualize it as sequence modeling (SM), effectively transforming extensive datasets into potent decision-making systems. 
This approach is advantageous for multi-task RL, offering a high-capacity model that accommodates task discrepancies and assimilates comprehensive knowledge from diverse datasets. 
However, the application of such high-capacity sequential models to multi-task RL introduces significant algorithmic challenges. 
In this section, we delineate the primary challenges, particularly conflicting gradients, and explore the concepts of parameter subspace and harmony, laying the groundwork for the motivation behind our method.

\subsection{Conflicting Gradients} 
\label{sec:conflict}

In a multi-task training context, the aggregate gradient, $\hat{\rvg}$, is computed across multiple tasks and is defined as
\begin{equation}
    \hat{\rvg}= \E_{\gT_i \sim p(\gT)} \nabla \gL_{\gT_i}(\theta)  =\frac{1}{N}\sum_{i=1}^N \rvg_i(\theta),
\end{equation}
where $\theta$ represents the trainable parameter vector and $\rvg_i$ is the gradient vector for task $\gT_i$.
In scenarios where tasks are diverse, the gradients $\rvg_i$ from different tasks may conflict significantly, a phenomenon known as gradient conflicts and negative transfer in multi-task learning \citep{guangyuan2022recon, tang2023concrete}.

\begin{definition}[Harmony Score on a Single Weight]
    The harmony score of the j-th element in the weights vector is estimated by calculating the corresponding coordinate of the element-wise product of the task gradient and the total gradient, denoted as $(\rvg_i \odot \hat{\rvg})_j=\rvg_{i,j} \times \hat{\rvg}_j$.
\end{definition}

\begin{definition}[Averaged Harmony Score]
    The overall harmony score across all weights is evaluated using $\frac{1}{NK}\sum_{i=1}^{N}\sum_{j=1}^K \frac{(\rvg_i \odot \hat{\rvg})_j}{|\rvg_{i,j}| |\hat{\rvg}_j|}$, where $K$ and $N$ denote the number of weights and tasks. This score, ranging between -1 and 1, reflects the degree of alignment among tasks.
\end{definition}

To substantiate the presence of conflicts in MTRL, we establish two metrics to measure harmony scores and conduct experiments utilizing the Prompt-DT method on 5 and 50 tasks from the Meta-world benchmark, recording the average harmony score.
As illustrated in Figure~\ref{fig:intro_4}, the averaged harmony score significantly diminishes with the escalation in the number of tasks, indicating pronounced conflicts among tasks and underscoring the imperative to address these conflicts in MTRL.

\subsection{Parameter Subspace and Harmony} 
\label{sec:redun}
Parameter subspace, a concept prevalent in pruning-aware training \citep{pruning}, aims to maintain comparable performance while achieving a sparse model. 
In the context of MTRL via SM, pruning to preserve distinct parameter subspaces for each task markedly alleviates gradient conflicts.
To validate this, we conduct experiments on 50 tasks from the Meta-World benchmark. 
Each task $\gT_i$ is assigned a randomly initialized mask $\mM^{\gT_i}$ with a specific sparsity ratio $\mathrm{S}$. 
During training, this mask modulates both the trainable parameters and gradients as follows:
\begin{equation}
\small
\label{eq:mask_gradient}
    \bar{\rvg}_i =  \nabla \gL_{\gT_i}(\theta \odot \mM^{\gT_i}) \odot \mM^{\gT_i}, ~~ i = 1, 2, \dots, N,
\end{equation}
where $\odot$ represents element-wise multiplication, and $\mM^{\gT_i}$ is a binary vector.
Intriguingly, as shown in Figure~\ref{fig:intro_5}, applying the mask could result in enhanced performance across a wide range of sparsity ratios. 
This improvement, coupled with a significantly higher harmony score in multi-task settings as depicted in Figure~\ref{fig:intro_4}, suggests that maintaining a subspace of parameters through the implementation of task-specific masks effectively mitigates the conflicts arising from unregulated parameter sharing.

\subsection{Motivation} 
The complexity in MTRL is significantly amplified with increasing task numbers, largely attributed to escalating gradient conflicts.
This challenge mainly stems from the unregulated sharing of parameters among tasks, intended to leverage similarities and enhance learning efficiency, but often leading to performance degradation.
In response to these challenges, a nuanced approach involving task-specific masks has been proposed. These masks aim to maintain distinct parameter subspaces for each task, thereby ensuring that the learning process of one task does not adversely affect the others. While this strategy represents a step towards mitigating gradient conflicts, it introduces a new challenge: the determination of an optimal mask configuration for each task remains an open question. The complexity of this challenge is compounded by the dynamic and often non-linear nature of task interactions within the shared model space.

Addressing this intricate problem requires a sophisticated solution that can navigate the delicate balance between shared learning and task-specific adaptation.
Our study proposes a bi-level optimization strategy situated within a meta-learning framework to address this issue. This approach leverages gradient-based techniques to meticulously explore and exploit the parameter space, aiming to identify a harmony parameter subspace for each task, thereby optimizing the overall MTRL performance.

\section{Method: Find Harmony Subspace}
\label{sec:method}

To address the aforementioned problem, we introduce a meta-learning framework that discerns an optimal harmony subspace of parameters for each task, enhancing parameter sharing and mitigating gradient conflicts. 
This problem is formulated as a bi-level optimization, where we meta-learn task-specific masks to define the harmony subspace. 
Mathematically, we can express the problem as:
\begin{align}
    \max_{\sM} ~ & ~ \E_{\gT_i \sim p(\gT)} [\sum_{t=0}^{\infty} \gamma^t \gR^{\gT_i}(s_t, \pi (\tau_{i,t}^{input} | \theta^{*\gT_i}))], \\
    \text{s.t.} ~~ & \theta^* = \argmin_{\theta} \E_{\gT_i \sim p(\gT)} \gL_{DT} (\theta, \sM), \\
    \text{where}& ~ \theta^{*\gT_i}=\theta^* \odot \mM^{\gT_i}, \sM = \{ \mM^{\gT_i} \}_{\gT_i \sim p(\gT)},
\end{align}
where $\mM^{\gT_i}$ represents a binary task mask vector corresponding to $\gT_i$, and $\sM$ denotes the set of all task masks.
The goal at the upper level is to learn a task-specific mask that identifies the harmony subspace for each task. 
Concurrently, at the inner level, the objective is to optimize the algorithmic parameters $\theta$, maximizing the collective performance of the unified model under the guidance of the task-specific masks.
The framework for our harmony subspace learning is depicted in Figure \ref{fig:pipeline}.
Subsequent sections are meticulously dedicated to elucidating the methodology for selecting the harmony subspace, detailing the sophisticated update mechanism for task masks (refer to Section \ref{sec:find}), and delineating the procedural intricacies of the algorithm (see Section \ref{sec:infer}).

\begin{algorithm}[t!]
    {\footnotesize
    \caption{HarmoDT}
    \label{alg:harmodt}  
    \begin{algorithmic}
    \STATE {\bfseries Input:} A set of tasks $\sT = \{ \gT_1,\dots,\gT_N \}$, maximum iteration $E$, episode length $T$, target return $\tG^*$, learning rate $\eta$, hyper-parameters $\{\eta_{\max}, \eta_{\min}, s, t_m, \lambda, \text{thresh} \}$.
    \STATE {\color{gray}\te{// initializing stage}} \\
    \STATE Initialize the parameters of the network $\theta_0$ and the set of task masks  $\sM = \{\mM^{\gT_1},\dots, \mM^{\gT_N}\}$ through ERK.
    \STATE {\color{gray}\te{// training stage}} \\
    \STATE $t = 1$.
    \WHILE{$t \leq E$}
        \STATE $\alpha_t=\lceil\eta_{max}+\frac{1}{2}(\eta_{min}-\eta_{max})(1+\cos{(2\pi \frac{t}{E})})\rceil$.
        \STATE $\sM$=Mask\_Update($\sT, \sM, \theta_t, \lambda, \alpha_t$).

        \WHILE{$t \bmod t_m \neq 0$}
            \STATE sample a task $\gT_i$ from the set of tasks $\sT$.
            \STATE sample a batch of data $\tau_i$ from the dataset $\gD_i$. %corresponding to task $\gT_i$.
            \STATE $\theta \leftarrow \theta - \eta \nabla \gL_{\gT_i}(\theta_t \odot \mM^{\gT_i}) \odot \mM^{\gT_i}$.
            \STATE $t = t + 1$.
        \ENDWHILE
        \STATE $t = t+1$.
    \ENDWHILE
    \STATE {\color{gray}\te{// inference stage}} \\
    \FOR{$i = 1, \dots, N$}
        \STATE Initialize history $\tau$ with zeros, desired reward $\hat{r} = \tG_i^*$, prompt $\tau^*$, the parameters $\theta_i \leftarrow \theta \odot \mM^{\gT_i}$.
        \STATE $j=0$.
        \FOR{$j \leq T$}
            \STATE Get action $a = \transformer_{\theta_i}(\tau^*, \tau)[-1]$.
            \STATE Step env. and get feedback $\rvs, \rva, r, \hat{r} \leftarrow \hat{r} - r$.
            \STATE Append $[\rvs, \rva, \hat{r}]$ to recent history $\tau$.
        \ENDFOR
    \ENDFOR
    \end{algorithmic}
 }
\end{algorithm}

\subsection{Mask Update}\label{sec:find}

For a given sparsity $\mathrm{S}$ and task masks $\sM$, we periodically assess the harmony subspace of trainable weights $\theta$ across all tasks. 
This process involves masking\footnote{When the term "mask" is used as a verb, it refers to the action of rendering the corresponding parameter inactive or modifying the mask vector by changing the value from 1 to 0.} $\alpha_t$ of the most conflicting parameters and subsequently recovering an equal number of previously masked parameters that have transitioned to harmony after the subsequent iterative training process. 
As delineated in Algorithm~\ref{alg:update}, this procedure encompasses three key steps: Weights Evaluation, Weights Masking, and Weights Recovery.

\paragraph{Weights Evaluation.}
During training, our aim is to iteratively identify a harmony subspace for each task by assessing trainable parameter conflicts and importance. 
This involves defining two metrics: the Agreement Score and the Importance Score, to gauge the concordance and significance of weights respectively.

\begin{definition}[Agreement Score]
For each task $\gT_i$ with a set of task masks $\sM$, the agreement score vector of all trainable weights is defined as follows: 
$A(\gT_i)=\bar{\rvg}_{i}\odot \frac{1}{N} \sum_{i=1}^N \bar{\rvg}_{i}$,
where $\bar{\rvg}_{i}$ is defined in Equation \ref{eq:mask_gradient}.
\end{definition}
\begin{definition}[Importance Score]
This metric evaluates the significance of parameters for task $\gT_i$. 
It can be measured either by the absolute value of the parameters $I_M(\gT_i)=|(\theta^{\gT_i})|$, indicating magnitude-based importance, or by the Fisher information $I_F(\gT_i)= \left(\nabla \log \gL_{\gT_i}\left( \theta^{\gT_i}\right) \odot \mM^{\gT_i} \right)^2$, reflecting the parameters' impact on output variability.
\end{definition}

For task $\gT_i$, $A(\gT_i)$ reflects the gradient similarity between the task-specific and the average masked gradients, while $I_M(\gT_i)_j$ and $I_F(\gT_i)_j$ measure the j-th element's importance. 
Lower values of $A(\gT_i)_j$ or $I_{M/F}(\gT_i)_j$ indicate increased conflict or diminished importance regarding the j-th element of the trainable parameters for the respective task.
The Harmony Score $H(\gT_i)_j$ for the j-th parameter of task $\gT_i$ is computed as a weighted balance between the Agreement and Importance Scores, moderated by a balance factor $\lambda$:
\begin{equation}
    \label{eq:init}
    \!\!H(\gT_i)_j=\left\{
    \begin{array}{ll}
    \!\!A(\gT_i)_j+\lambda I_{M/F}(\gT_i)_j, & (\mM^{\gT_i})_j =1, \\
    \!\! \inf ,  & (\mM^{\gT_i})_j =0.\\
    \end{array} \right. 
\end{equation}
Parameters that have been already masked (i.e., $(\mM^{\gT_i})_j = 0$) are assigned an infinite Harmony Score to prevent their re-selection due to the pre-existing conflicts.

\begin{algorithm}[t!]
    {\footnotesize
    \caption{Mask Update}
    \label{alg:update}
    \begin{algorithmic}
    \STATE {\bfseries Input:} A set of tasks $\sT$, a set of task masks $\sM$, trainable weights vector $\theta_t$ and hyper-parameters $\{ \lambda, \alpha_t \}$.
    \FOR{$i= 1,\dots,N$}
        \STATE $\bar{\rvg}_i= \nabla \gL_{T_i} (\theta \odot \mM^{\gT_i}) \odot \mM^{\gT_i}$.
        \STATE $\rvg_i= \nabla \gL_{T_i} (\theta) $.
    \ENDFOR
    \STATE $\hat{\rvg}=\frac{1}{N} \sum_{i=1}^N \bar{\rvg}_i$.
    \FOR{$i= 1,\dots,N$}
        \STATE {\color{gray}\te{// Weights Evaluation}} \\
        \STATE Calculate $H(\gT_i)$ with $\lambda$, $\rvg_i$ and $\hat{\rvg}$ as Sec.~\ref{sec:find}.
        \STATE {\color{gray}\te{// Weights Masking}} \\
        \STATE $\mM^{\gT_i}=\mM^{\gT_i}-\operatorname{ArgBtmK}_{\alpha_t}\left(H(\gT_i)\right).$
        \STATE {\color{gray}\te{// Weights Recovery}}
        \STATE $\mM^{\gT_i}=\mM^{\gT_i}+\operatorname{ArgTopK}_{\alpha_t}(\rvg_i \odot \hat{\rvg}).$
    \ENDFOR
    \end{algorithmic}
{\textbf{Output}:$\sM.$}
}
\end{algorithm}

\paragraph{Weights Masking.}
Employing the Harmony Score, we identify and mask the most conflicting and less significant weights within the harmony subspace as follows:
\begin{equation}
\label{eq:weightmask}
    \mM^{\gT_i}=\mM^{\gT_i}-\operatorname{ArgBtmK}_{\alpha_t}\left(H(\gT_i)\right),
\end{equation}
where $\alpha_t$ represents the number of masks altered in the t-th iteration, and $\operatorname{ArgBtmK}_{\alpha_t}(\cdot)$ generates zero vectors matching the dimension of $\mM^{\gT_i}$, marking the positions of the top-$\alpha_t$ smallest values from $H(\gT_i)$ with 1.

\paragraph{Weights Recover.}
To maintain a fixed sparsity ratio and recover weights that have transitioned from conflict to harmony, the following recovery process is applied:

\begin{equation}
\mM^{\gT_i}=\mM^{\gT_i}+\operatorname{ArgTopK}_{\alpha_t}(\rvg_i \odot \frac{1}{N}\sum_{i=1}^N \bar{\rvg}_i),
\end{equation}
where $\operatorname{ArgTopK}_{\alpha_t}(\cdot)$ generates zero vectors matching the dimension of $\mM^{\gT_i}$, setting the positions corresponding to the top-$\alpha_t$ largest values from $(\rvg_i \odot \frac{1}{N}\sum_{i=1}^N \bar{\rvg}_i)$ to 1.
This step ensures the reintegration of previously conflicting weights that have harmonized after subsequent iterations.

\subsection{Training and Inference}
\label{sec:infer}

Algorithm~\ref{alg:harmodt} provides the meta-learning process for the task mask set $\sM$ and the update mechanism for the trainable parameters $\theta$ of the unified model.
Given a set of source tasks, we first initialize corresponding masks through the ERK technique \citep{ERK} with a predefined sparsity ratio $\mathrm{S}$ for each task (See Appendix~\ref{app:ERK} for more details).
In the inner loop, we optimize the parameters of the unified model under the guidance of the task-specific mask:
\begin{equation}
    \small
    \theta_{t+1}=\theta_{t}-\eta \E_{\gT_i \sim p(\gT)} \nabla \gL_{\gT_i}(\theta \odot \mM^{\gT_i}) \odot \mM^{\gT_i}.
\end{equation}
Then, in the outer loop, we optimize the set of task masks through the procedure detailed in Section \ref{sec:find} to find the harmony subspace for each separate task.
Considering the stability and efficiency of the updating, we adopt a warm-up and cosine annealing strategy~\cite{cos} to control the updating number $\alpha_t$. Given the maximum iterations $E$, $\alpha_t$ in t-th iteration is defined as:
\begin{equation}
    \alpha_t=\lceil\eta_{max}+\frac{1}{2}(\eta_{min}-\eta_{max})(1+\cos{(2\pi \frac{t}{E})})\rceil,
\end{equation}
where $\lceil \cdot \rceil$ represents the round-up command, and $\eta_{min}$ and $\eta_{max}$ denote the lower and upper bounds, respectively, on the number of parameters that undergo changes during the mask update process.

In the inference stage, task IDs are accessible in the online test environment, following the methodology of \citet{he2023diffusion}.
Accordingly, the task-specific mask is applied to the parameters, and the inference process is conducted in line with \citet{PDT}. 
For unseen tasks that differ from training tasks but share identical states and transitions, we aggregate task-specific masks from training tasks to formulate a generalized model. 
The mask for unseen tasks is constructed as follows:
\begin{equation}
    \label{eq:unseen}
    \hat{M}_j=\left\{
    \begin{array}{ll}
0, & \sum_{i=1}^N M_{i,j} \leq \text{thresh},\\
1,  & \sum_{i=1}^N M_{i,j} > \text{thresh},\\
\end{array} \right. 
\end{equation}
where $\hat{M}$ denotes the mask for the unseen task, and `thresh' is a predefined threshold. 
This approach integrates insights from all training tasks, enhancing the model's adaptability to novel scenarios.
\section{Experiment}
\label{sec:exp}

In this section, we conduct extensive experiments to answer the following questions:
(1) How does HarmoDT compare to other offline and online baselines in the multi-task regime?
(2) Does HarmoDT mitigate the phenomenon of conflicting gradients and identify an optimal harmony subspace of parameters for each task?
(3) Can HarmoDT\footnote{Our code is available at: \url{https://github.com/charleshsc/HarmoDT}} generalize to unseen tasks?

\subsection{Environments and Baselines}

\paragraph{Environment.}
Our experiments utilize the Meta-World benchmark \citep{yu2020meta}, featuring 50 distinct manipulation tasks with shared dynamics, requiring a Sawyer robot to interact with various objects. We extend tasks to a random-goal setting, consistent with recent studies \citep{he2023not, sun2022paco}.
Performance evaluation is based on the averaged success rate across tasks. 
Following \citet{he2023diffusion}, we employ two dataset compositions: a near-optimal dataset from SAC-Replay \citep{sac} ranging from random to expert experiences, and a sub-optimal dataset with initial trajectories and a reduced proportion (50\%) of expert data.

For unseen tasks, HarmoDT's performance is evaluated on distinct objectives from datasets used in prior works \citep{mitchell2021offline, yu2020meta, PDT, PTDT}, specifically Cheetah-dir, Cheetah-vel, and Ant-dir, which challenge the agent to optimize direction and speed. 
Details on environment specifications and hyper-parameters are available in the Appendix \ref{sec:env} and \ref{sec:hypar}.

\begin{table}[t]
\setlength{\tabcolsep}{3pt}
\centering
\caption{Average success rate across 3 seeds on Meta-World MT50 with random goals (MT50-rand) under both near-optimal and sub-optimal cases.
Each task is evaluated for 50 episodes.
Approaches with * indicate baselines of our own implementation.
}
\vspace{0.1cm}
\small
\centering
\scalebox{1.00}{
\begin{tabular}{l|ccc}
\toprule[2pt]
Method& \multicolumn{3}{c}{ Meta-World 50 Tasks} \\
\cmidrule{1-4} \#Partition &Near-optimal  & Sub-optimal & Params\\
\midrule 
\textbf{CARE}~(online) & $46.12_{\pm 1.30}$ & \textbf{-}&1.26 M \\
\textbf{PaCo}~(online) & $54.31_{\pm 1.32}$ & \textbf{-}&3.39 M\\
\textbf{Soft-M}~(online) & $53.41_{\pm 0.72}$ & \textbf{-}& 1.62 M\\
\textbf{D2R}~(online) & $63.53_{\pm 1.22}$ & \textbf{-}& 1.40 M\\
\midrule
\textbf{MTBC}  & $60.39_{\pm 0.86}$ & $34.53_{\pm 1.25}$&1.74 M\\
\textbf{MTIQL} & $56.21_{\pm 1.39}$ & $43.28_{\pm 0.90}$&1.74 M \\
\textbf{MTDIFF-P} & $59.53_{\pm 1.12}$ & $48.67_{\pm 1.32}$&5.32 M \\
\textbf{MTDIFF-P-ONEHOT}& $61.32_{\pm 0.89}$ & $\textbf{48.94}_{\pm 0.95}$&5.32 M \\
% \midrule
\textbf{MTDT} & $20.99_{\pm 2.66}$ & $20.63_{\pm 2.21}$& 0.87 M\\
\textbf{MTDT*} & $ 65.80_{\pm 1.02}$ & $42.33_{\pm 1.89}$& 1.47 M \\
\textbf{Prompt-DT} & $45.68_{\pm 1.84}$ & $39.76_{\pm 2.79}$ &0.87 M\\
\textbf{Prompt-DT*} & $\textbf{69.33}_{\pm 0.89}$ & $48.40_{\pm 0.16}$&1.47 M \\
\midrule
\textbf{HarmoDT-R~(ours)} & $\textbf{75.39}_{\pm 1.18}$ & $\textbf{53.80}_{\pm 1.07}$&1.47 M \\
\textbf{HarmoDT-M(ours)} & $\textbf{80.33}_{\pm 0.97}$ & $\textbf{57.20}_{\pm 0.73}$ &1.47 M\\
\textbf{HarmoDT-F(ours)} & $\textbf{82.20}_{\pm 0.40}$ & $\textbf{57.20}_{\pm 0.68}$ &1.47 M\\
\bottomrule[2pt]
\end{tabular}
}
\label{tb:50task}
\vspace{-0.5cm}
\end{table}

\begin{table*}[t!]
\centering
\caption{We randomly select 5, 30, 50 tasks from the Meta-World benchmark under both near-optimal and sub-optimal cases and record the average success rate across 3 seeds. Each task is evaluated for 50 episodes.}
\vspace{2pt} 
\centering
\scalebox{1.00}{
\begin{tabular}{l|cc|cc|cc}
\toprule[2pt]
Method& \multicolumn{2}{c|}{ Meta-World 5 Tasks} & \multicolumn{2}{c|}{ Meta-World 30 Tasks} & \multicolumn{2}{c}{Meta-World 50 Tasks} \\
\cmidrule{1-7} \#Partition &Near-optimal  & Sub-optimal &  Near-optimal   & Sub-optimal&  Near-optimal   & Sub-optimal\\
\midrule 
\textbf{MTDIFF} & $\textbf{100.0}\pm 0.0$ & $66.30\pm 2.31$& $67.52\pm 0.35$ & $\textbf{54.21}\pm 1.10$ & $61.32\pm 0.89$ & $\textbf{48.94}\pm 0.95$ \\
\textbf{MTDT} & $\textbf{100.0} \pm 0.0$ & $64.67\pm 5.25$ & $ 71.89 \pm 0.95$ & $ 49.33\pm 2.05$ & $ 65.80\pm 1.02$ & $42.33\pm 1.89$ \\
\textbf{Prompt-DT} & $\textbf{100.0}\pm 0.0$ & $\textbf{66.50}\pm 2.74$ & $\textbf{74.10}\pm 0.69$ & $53.78\pm 0.63$& $\textbf{69.33}\pm 0.89$ & $48.40\pm 0.16$ \\
\midrule
\textbf{HarmoDT-R} & $\textbf{100.0}\pm 0.0$ & $66.14\pm 3.00$& $82.65\pm 2.11$ & $61.22\pm 2.16$& $75.39\pm 1.18$ & $53.80\pm 1.07$ \\
\textbf{HarmoDT-M} & $\textbf{100.0}\pm 0.0$ & $74.00\pm 4.79$& $84.67\pm 0.84$ & $63.67\pm 0.73$& $80.33\pm 0.97$ & \textbf{$57.20\pm 0.73$} \\
\textbf{HarmoDT-F} & $\textbf{100.0}\pm 0.0$ & $\textbf{77.67}\pm 3.25$ & $\textbf{88.00}\pm 0.96$ & $\textbf{65.00}\pm 0.92$ & $\textbf{82.20}\pm 0.40$ & $\textbf{57.20}\pm 0.68$ \\
\bottomrule[2pt]
\end{tabular}
}
\label{tb:scale}
    \vspace{-0.3cm} 
\end{table*}

\begin{figure*}[t!]
    \centering
    \subfigure{
    \centering
    \label{fig:s3}
    \includegraphics[width=0.23\textwidth]{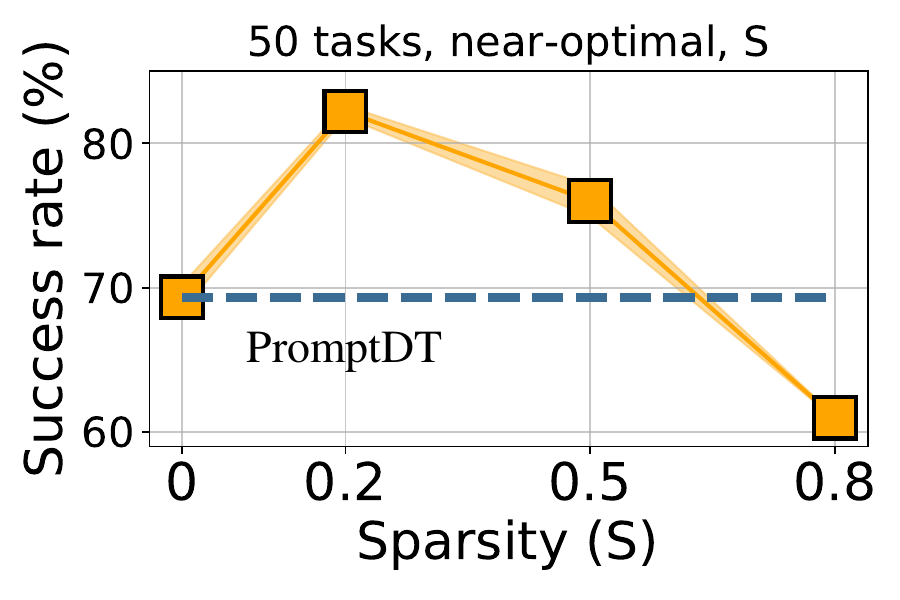}}
    \centering
    \subfigure{
    \centering
    \label{fig:e3}
    \includegraphics[width=0.23\textwidth]{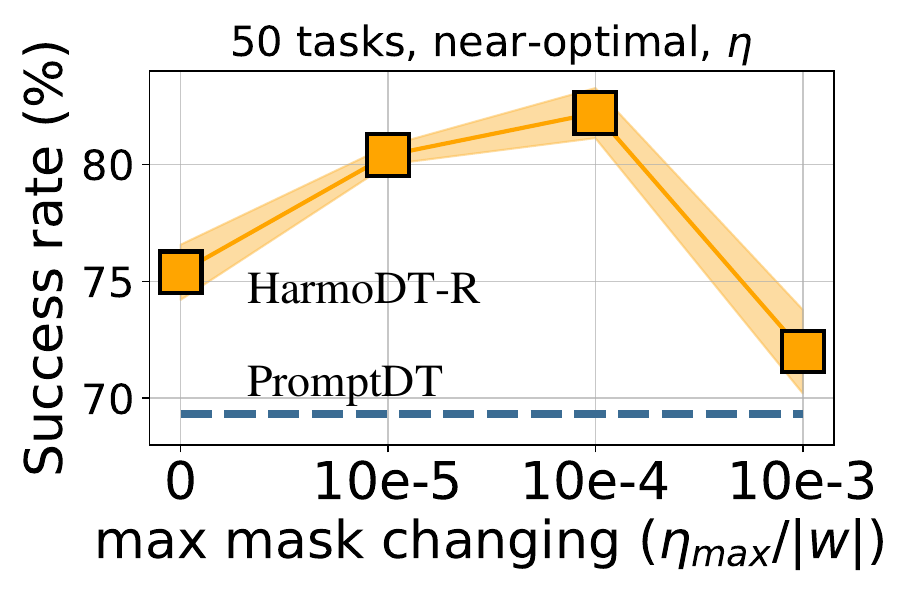}}
    \subfigure{
    \centering
    \label{fig:l3}
    \includegraphics[width=0.23\textwidth]{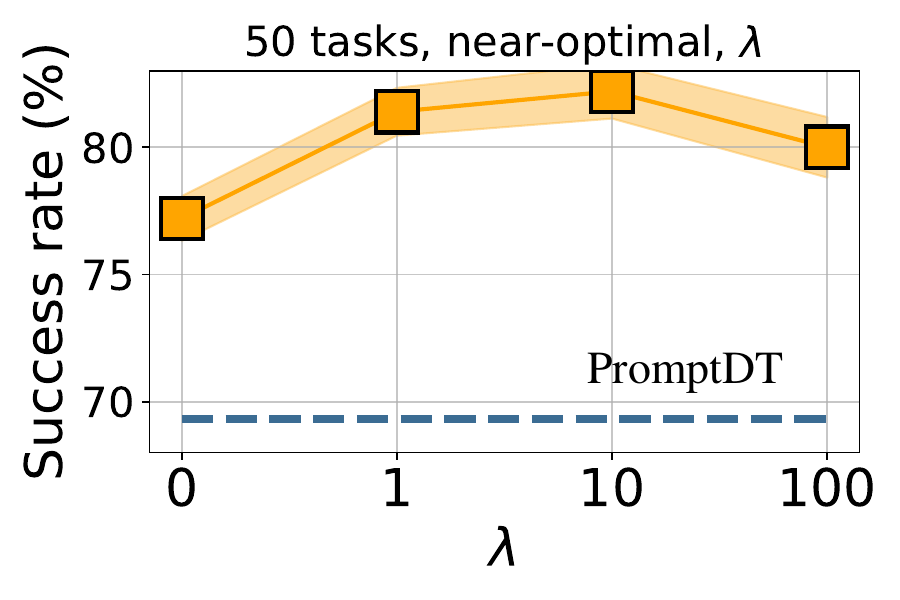}}
    \subfigure{
    \centering
    \label{fig:t3}
    \includegraphics[width=0.23\textwidth]{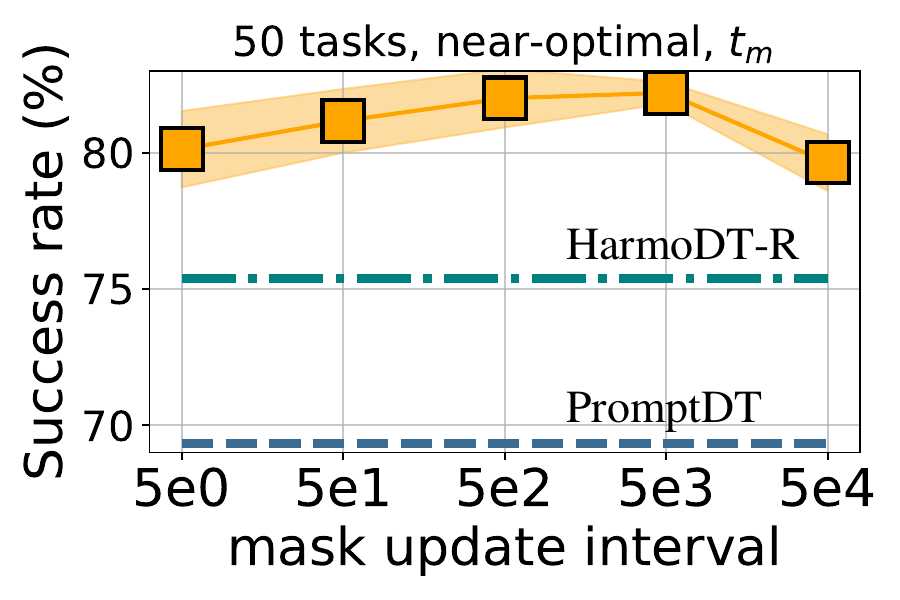}}
    \vspace{-0.2cm}
    \caption{From the left to right, we illustrate the ablation results on the Meta-World benchmark with 50 tasks under the near-optimal case. 
    Default values are listed as $\eta_{min}=0$, $\eta_{max}$ is 100~(about 1e-3\% of total weights), $\mathrm{S}=0.2$, $\lambda=10$ and $t_m=5e3$.
    During each individual ablation, a single parameter is varied, with all other parameters maintained at their default values. Detailed results pertaining to additional settings are comprehensively documented in the Appendix \ref{sec:moreab}.}
    \label{fig:ablation}
        \vspace{-0.3cm}
\end{figure*}

\paragraph{Baselines.}
We compare our proposed HarmoDT with the following offline baselines.
(i) \textbf{MTBC}: Extends Behavior Cloning for multi-task learning with network scaling and a task-ID conditioned actor;
(ii) \textbf{MTIQL}: Adapts IQL \citep{IQL} with multi-head critics and a task-ID conditioned actor for multi-task policy learning;
(iii) \textbf{MTDIFF-P} \citep{he2023diffusion}: A diffusion-based method combining Transformer architectures and prompt learning for generative planning in multitask offline settings;
(iv) \textbf{MTDT}: Extends DT \citep{DT} to multitask settings, utilizing a task ID encoding and state input for task-specific learning;
(v) \textbf{Prompt-DT} \citep{PDT}: Builds on DT, leveraging trajectory prompts and reward-to-go for multi-task learning and generalization to unseen tasks.
The results of these offline methods are directly replicated from \citet{he2023diffusion}, with the exception that the approaches marked with * are implemented by us.

Besides, we compare our method with four online RL methods: 
(vi) \textbf{CARE} \citep{sodhani2021multi}: Utilizes metadata and a mixture of encoders for task representation;
(vii) \textbf{PaCo} \citep{sun2022paco}: Employs a parameter compositional approach for task-specific parameter recombination;
(viii) \textbf{Soft-M} \citep{yang2020multi}: Focuses on a routing network for the soft combination of modules;
(ix) \textbf{D2R} \citep{he2023not}: Adopts disparate routing paths for module selection per task.
The results of these methods are directly extracted from \citet{he2023not}.
Detailed descriptions of these baselines are summarized in the Appendix~\ref{app:baseline}.

\subsection{Main Results} 
In this study, we benchmark HarmoDT and its variants against established baselines on 50 Meta-World tasks.
The variants considered in this evaluation include \textbf{HarmoDT-R}, which maintains frozen task masks throughout the training process; 
\textbf{HarmoDT-F}, which utilizes fisher information $I_F(\gT_i)$ for calculating weight importance; 
and \textbf{HarmoDT-M}, which employs magnitude information $I_M(\gT_i)$ for determining weight importance.
Note that the primary distinction between HarmoDT-F and HarmoDT-M resides in their respective approaches to weight masking (Equation \ref{eq:weightmask}).

As shown in Table~\ref{tb:50task}, HarmoDT-R, following Prompt-DT's structure, surpasses all other methods, achieving a 6.1\% and 4.9\% improvement in near-optimal and sub-optimal scenarios, respectively, compared to the best baseline. 
By employing fixed random masks, HarmoDT-R effectively competes with the current state-of-the-art techniques.
Furthermore, the variants, HarmoDT-M and HarmoDT-F, enhance the performance of HarmoDT-R by identifying optimal harmony subspaces through task mask learning, resulting in substantial gains of 6.8\% and 3.4\% in near-optimal and sub-optimal cases, respectively.
Our novel technique, HarmoDT, showcases its effectiveness in multi-task settings, encompassing both sub-optimal datasets that require the stitching of useful segments from suboptimal trajectories and near-optimal datasets where the emulation of optimal behaviors is crucial.

\subsection{Further Analysis}

This section delves into the scalability of task scale, model size, and hyper-parameters. It also examines the visualization of task masks for the harmony subspace and evaluates the performance on unseen tasks.

\textbf{Scalability of Task Scale.} 
We evaluated HarmoDT's scalability across varying task numbers in the Meta-World benchmark. 
As shown in Table~\ref{tb:scale} and Figure \ref{fig:performance_num}, HarmoDT consistently outperforms MTDIFF, MTDT, and Prompt-DT across all task numbers and cases, demonstrating promising superiority with increasing task count: 11\% for 5 tasks, 11\% for 30 tasks, and 8\% for 50 tasks in sub-optimal settings.

\begin{table}[t!]
\setlength{\tabcolsep}{4pt}
\centering
\caption{Ablation study on the model size of MTDT, Prompt-DT, and our HarmoDT-F under near-optimal of Meta-World 30 tasks and 50 tasks. We denote the model with z M parameters and x layers of y head attentions as (x, y, z) in the table.}
\vspace{2pt}
\label{tab:modelsize}
\small
\centering
\scalebox{1.0}{
\begin{tabular}{c|c|ccc}
\toprule[2pt]
 Case & Model & MTDT & Prompt-DT & HarmoDT-F \\
\midrule 
\multirow{4}{*}[0pt]{30 tasks} 
 & (1, 2, 0.48) & $35.78_{\pm 0.9}$ & $42.10_{\pm 3.1}$& $59.33_{\pm 1.5}$\\
 & (3, 4, 0.87) & $64.01_{\pm 1.4}$ & $69.56_{\pm 1.0}$ & $77.67_{\pm 1.6}$\\
 & (6, 8, 1.47) & $71.89_{\pm 1.0}$ & $74.10_{\pm 0.7}$& $88.00_{\pm 1.0}$ \\
 & (9, 16, 2.06) & $74.33_{\pm 0.7}$ & $77.67_{\pm 1.7}$& $88.00_{\pm 1.6}$ \\
\midrule
  \multirow{4}{*}[0pt]{50 tasks} 
 & (1, 2, 0.48) & $31.93_{\pm 3.4}$ & $36.07_{\pm 2.5}$& $58.20_{\pm 1.9}$\\
 & (3, 4, 0.87) & $58.13_{\pm 3.6}$ & $61.93_{\pm 0.4}$ & $72.80_{\pm 1.6}$\\
 & (6, 8, 1.47) & $65.80_{\pm 1.0}$ & $69.33_{\pm 0.9}$& $82.20_{\pm 0.4}$ \\
 & (9, 16, 2.06) & $68.33_{\pm 1.1}$ & $71.00_{\pm 0.9}$& $83.33_{\pm 0.6}$ \\
\bottomrule[2pt]
\end{tabular}
}
\vspace{-0.4cm}
\end{table}

\begin{table}[t!]
\setlength{\tabcolsep}{4pt}
\centering
\caption{Generalization ability to unseen tasks. Here we conduct experiments and record the cumulative reward of unseen tasks on three distinct datasets: Cheetah-dir, Cheetah-vel, and Ant-dir, which challenge the agent to optimize direction and speed.}
\vspace{2pt}
\label{tb:unseen}
\small
\centering
\scalebox{1.0}{
\begin{tabular}{L{1.8cm}|C{1.7cm}C{1.7cm}C{1.7cm}}
\toprule[2pt]
Setting & MTDT & Prompt-DT & HarmoDT-F \\
\midrule 
Cheetah-dir & $662.40_{\pm 1.3}$ & $935.3_{\pm 2.6}$& $\textbf{958.5}_{\pm 1.5}$ \\
Cheetah-vel & $-170.11_{\pm 5.7}$ & $-127.7_{\pm 9.9}$& $\textbf{-66.51}_{\pm 1.2}$\\
Ant-dir & $165.29_{\pm 0.4}$ & $278.7_{\pm 38.7}$& $\textbf{298.3}_{\pm 1.0}$ \\
\midrule 
Average & $219.2$ & $362.1$& $\textbf{396.8}_{9.6\% \uparrow}$ \\
\bottomrule[2pt]
\end{tabular}
}
\vspace{-0.4cm}
\end{table}

\textbf{Impact of Model Size.} 
The influence of model size is pronounced in multi-task training scenarios. 
Table \ref{tab:modelsize} delineates our ablation study on model size across 1e6 iterations. Models are characterized by their parameters (z M), layers (x), and head attentions (y), represented as (x, y, z).
Results reveal that increasing model size markedly boosts performance for all evaluated methods. 
Significantly, our approach, HarmoDT, demonstrates consistent superiority over MTDT and Prompt-DT across a range of model sizes.
This success is attributed to HarmoDT's effective establishment of a harmony subspace for each task.

\textbf{Ablation on Hyper-parameters.} 
This study introduces hyper-parameters $\eta_{max}$ for cosine annealing (with $\eta_{min}=0$), mask alteration frequency $t_m$, overall sparsity $\mathrm{S}$, and balance controller $\lambda$.
Comprehensive ablations are conducted to establish an empirical strategy for selecting these parameters.
Figure~\ref{fig:ablation} delineates the ablation study on 50 tasks of Meta-World benchmark in the near-optimal settings. 
Ablation results on other settings can be found in the Appendix \ref{sec:moreab}.
Across a broad spectrum of hyper-parameter values, our approach consistently outperforms baselines. 
Based on these insights, recommended settings for hyper-parameters are sparsity ratio $\mathrm{S}=0.2$, $\eta_{max}$ at approximately 0.001\% of total weights, balance factor $\lambda=10$ and mask changing interval $t_m=5000$ rounds. 
These parameters collectively contribute to the superior performance of our method.

\begin{figure}[ht!]
\label{fig:intro_1}
\includegraphics[width=0.48\textwidth]{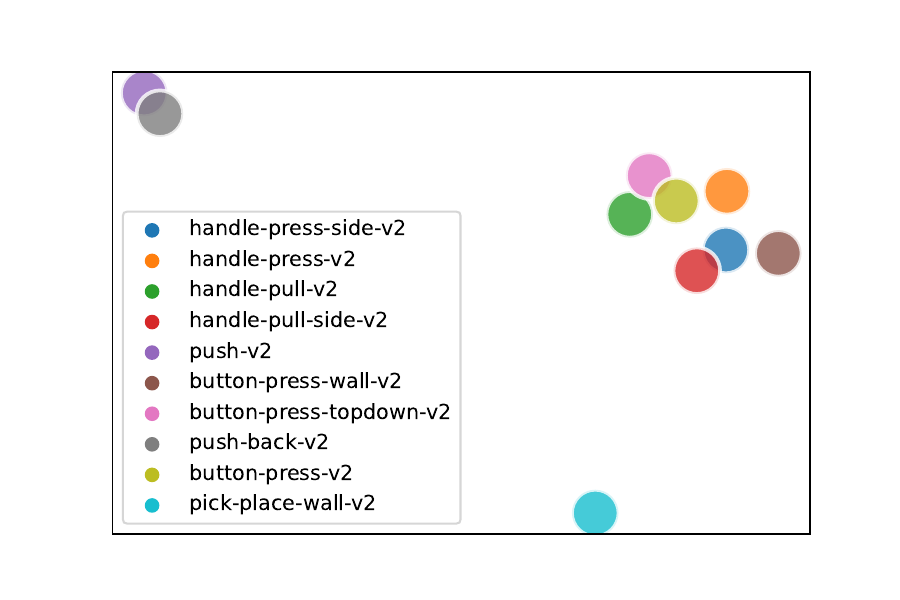}
\vspace{-0.6cm}
\caption{The t-SNE visualization of optimal subspace via masks learned by our HarmoDT on the 30 tasks of Meta-World benchmark. 
The figure illustrates the relational dynamics of task-specific masks, with a focus on 10 representative tasks from the total set. }
\label{fig:tsne}
\vspace{-0.1cm}
\end{figure}

\textbf{Visualization of Mask.}
As shown in Figure~\ref{fig:tsne}, we use t-SNE~\citep{tsne} to visualize the task masks post-training on 30 tasks from the Meta-World benchmark.
Note that even small distances in the visualization can represent significant divergences in the original high-dimensional parameter space.
The visualization effectively showcases the relational dynamics of the task masks; closely related tasks such as `push-back-v2' and `push-v2' are positioned in proximity, while disparate tasks like `push-v2' and `pick-place-wall-v2' are distinctly separated.
This spatial arrangement underscores the efficacy of our HarmoDT in delineating a harmony subspace tailored for each task.

\textbf{Ability to unseen tasks.} 
Prompt-DT's proficiency with unseen tasks prompted us to assess HarmoDT's capabilities in similar scenarios. 
We employ a voting mechanism among all observed tasks to define a generalized subspace for unseen tasks, as delineated in Equation \ref{eq:unseen}. 
This technique operates on a foundational assumption: parameters that are consistently identified as significant and harmonious across a range of tasks (surpassing a predefined threshold) are posited to hold universal value, potentially contributing positively to task performance in novel scenarios. 
Comparative analysis involving HarmoDT, MTDT, and Prompt-DT is conducted on three distinct datasets: Cheetah-dir, Cheetah-vel, and Ant-dir.
The results, presented in Table~\ref{tb:unseen}, affirm HarmoDT's comprehensive enhancements across all test cases. 
Notably, HarmoDT demonstrates an average reward of 396.8, surpassing Prompt-DT's 362.1 with a substantial 9.6\% improvement. 
This outcome underscores the efficacy of our voting approach in addressing unseen tasks.

\section{Conclusion}

In this study, we introduce the Harmony Multi-Task Decision Transformer (HarmoDT), a novel approach designed to discern an optimal parameter subspace for each task, leveraging parameter sharing to harness task similarities while concurrently addressing the adverse impacts of conflicting gradients.
By employing a bi-level optimization and a meta-learning framework, HarmoDT not only excels as a comprehensive policy in multi-task environments but also exhibits robust generalization capabilities to unseen tasks. 
Our rigorous empirical evaluations across a diverse array of benchmarks underscore HarmoDT's superior performance compared to existing baselines, establishing its state-of-the-art effectiveness in MTRL scenarios.

\textbf{Limitation. ~~} We present an innovative approach to policy learning in multi-task offline RL, achieving state-of-the-art performance across various tasks. However, the efficacy of our approach depends on the model's capacity, as it employs a sparsification strategy for each task's parameter space. Moreover, tasks of differing complexity inherently require varying numbers of parameters for optimal performance. Our method, however, uses the same number of task-specific parameters across different tasks. Complex tasks could benefit from a denser parameter allocation, while simpler tasks might achieve peak efficiency with a sparser configuration.

\section*{Acknowledgements}
This work is supported by STI 2030—Major Projects (No. 2021ZD0201405), STCSM (No. 22511106101, No. 22511105700, No. 21DZ1100100), 111 plan (No. BP0719010) and National Natural Science Foundation of China (No. 62306178). Dr Tao's research is partially supported by NTU RSR and Start Up Grants.

\section*{Impact Statement}
In domains such as healthcare and robotics, multi-task scenarios are commonplace. Our algorithm enhances the applicability of DT by extending them to complex multi-task environments, increasing their practical utility. However, the performance on certain tasks may deteriorate if they are provided with bad task masks, potentially due to malicious intent. Therefore, it's imperative to rigorously follow the algorithmic process during updates, treating each task with equal consideration. To date, our investigations have not uncovered any adverse societal impacts.

\bibliography{example_paper}
\bibliographystyle{icml2024}

%%%%%%%%%%%%%%%%%%%%%%%%%%%%%%%%%%%%%%%%%%%%%%%%%%%%%%%%%%%%%%%%%%%%%%%%%%%%%%%
%%%%%%%%%%%%%%%%%%%%%%%%%%%%%%%%%%%%%%%%%%%%%%%%%%%%%%%%%%%%%%%%%%%%%%%%%%%%%%%
% APPENDIX
%%%%%%%%%%%%%%%%%%%%%%%%%%%%%%%%%%%%%%%%%%%%%%%%%%%%%%%%%%%%%%%%%%%%%%%%%%%%%%%
%%%%%%%%%%%%%%%%%%%%%%%%%%%%%%%%%%%%%%%%%%%%%%%%%%%%%%%%%%%%%%%%%%%%%%%%%%%%%%%
\newpage
\appendix
\onecolumn

\section{Detailed Environment}
\label{sec:env}
\subsection{Meta-World}
The Meta-World benchmark, introduced by \citet{yu2020meta}, encompasses a diverse array of 50 distinct manipulation tasks, unified by shared dynamics. These tasks involve a Sawyer robot engaging with a variety of objects, each distinguished by unique shapes, joints, and connective properties. The complexity of this benchmark lies in the heterogeneity of the state spaces and reward functions across tasks, as the robot is required to manipulate different objects towards varying objectives. The robot operates with a 4-dimensional fine-grained action input at each timestep, which controls the 3D positional movements of its end effector and modulates the gripper's openness.
In its original configuration, the Meta-World environment is set with fixed goals, a format that somewhat limits the scope and realism of robotic learning applications. To address this and align with recent advancements in the field, as noted in works by \citet{sun2022paco, yang2020multi}, we have modified all tasks to incorporate a random-goal setting, henceforth referred to as MT50-rand. The primary metric for evaluating performance in this enhanced setup is the average success rate across all tasks, providing a comprehensive measure of the robotic system's adaptability and proficiency in varied task environments.

For the creation of the offline dataset, we follow the work by \citet{he2023diffusion} and employ the Soft Actor-Critic (SAC) algorithm \citep{sac} to train distinct policies for each task until they reach a state of convergence. Subsequently, we compile a dataset comprising 1 million transitions per task, extracted from the SAC replay buffer. These transitions represent samples observed throughout the training period, up until the point where each policy's performance stabilized. Within this benchmark, we have curated two distinct dataset compositions:
\begin{itemize}[leftmargin=*]
    \item \textbf{Near-optimal} dataset consisting of the experience (100M transitions) from random to expert (convergence) in SAC-Replay.
    \item \textbf{Sub-optimal} dataset consisting of the initial 50\% of the trajectories (50M transitions) of the near-optimal dataset for each task, where the proportion of expert data decreases a lot.
\end{itemize}

\subsection{Unseen Tasks}
In our evaluation, we apply our approach to a diverse array of meta-RL control tasks, each offering distinct challenges to assess the performance and generalization capabilities of our model. The tasks are detailed as follows:
\begin{itemize}[leftmargin=*]
    \item \textbf{Cheetah-dir}: This task involves two distinct directions: forward and backward. The objective is for the cheetah agent to achieve high velocity in the assigned direction. The evaluation encompasses both training and testing sets, covering these two directions comprehensively to gauge the agent’s performance effectively.
    \item \textbf{Cheetah-vel}: Here, the task defines 40 unique sub-tasks, each associated with a specific goal velocity, uniformly distributed between 0 and 3 m/s. The agent's performance is assessed based on the $l_2$ error relative to the target velocity, with a penalty for deviations. For testing, 5 of these tasks are selected, while the remaining 35 are used for training purposes.
    \item \textbf{Ant-dir}: This task comprises 50 different sub-tasks, each with a goal direction uniformly sampled in a two-dimensional plane. The agent, an 8-jointed ant, is incentivized to attain high velocity in the designated direction. Of these, 5 tasks are earmarked for testing, with the rest allocated for training.
\end{itemize}

By evaluating our approach on these diverse tasks, we can assess its performance and generalization capabilities across different control scenarios.
The generalization ability of our approach is rigorously tested by examining the distribution of tasks between the training and testing sets, as outlined in Table \ref{tab:set_meta}. This experimental setup, as described in Section \ref{sec:exp}, adheres to the divisions specified, ensuring consistency in our evaluation and facilitating a thorough assessment of our approach’s adaptability and effectiveness across varied control tasks.

\section{Hyper-parameters and Resources}
\label{sec:hypar}
This section elaborates on the specifics of the training regimen implemented in our study. During the training phase, tasks are randomly selected for model refinement. The configuration for each training iteration is meticulously set, with a batch size of 8 and the utilization of the Adam optimizer, operating at a learning rate of 1e-4. The total number of training steps is established at 10 million.
We build our policy as a Transformer-based model, which is based on minGPT open-source code \footnote{\url{https://github.com/karpathy/minGPT}}.
The specific model parameters and hyper-parameters utilized in our training process are outlined in Table \ref{tab:model_parameter}.

\begin{table}[h]
    \caption{Training and testing task indexes when testing the generalization ability in meta-RL tasks }
    \vspace{0.1cm}
    \label{tab:set_meta}
    \centering
    \begin{tabular}{ll}
      \toprule
      \multicolumn{2}{c}{Cheetah-dir} \\
      \midrule
      Training set of size 2 & [0,1] \\
      Testing set of size 2 & [0.1]\\
      \midrule
      \multicolumn{2}{c}{Cheetah-vel} \\
      \midrule
      Training set of size 35 & [0-1,3-6,8-14,16-22,24-25,27-39]\\
      Testing set of size 5 & [2,7,15,23,26]\\
      \midrule
      \multicolumn{2}{c}{Ant-dir} \\
      \midrule
      Training set of size 45 &  [0-5,7-16,18-22,24-29,31-40,42-49]\\
      Testing set of size 5 &  [6,17,23,30,41]\\
      \bottomrule
    \end{tabular}
    \vspace{-0.2in}
\end{table}

\begin{table}[h]
\renewcommand{\arraystretch}{1.3}
\centering
  \caption{Hyper-parameters of HarmoDT in our experiments.}
  \vspace{0.1cm}
  \label{tab:model_parameter}
  \begin{tabular}{ll}
    \hline
    Parameter & Value \\
    \hline
    Number of layers            & 6 \\
    Number of attention heads   & 8 \\
    Embedding dimension         & 256 \\
    Nonlinearity function       & ReLU \\
    Batch size                  & 8 \\
    Context length $K$          & 20 \\
    Dropout                     & 0.1 \\
    Learning rate               & 1.0e-4 \\
    Total rounds               & 1e6 \\
    Sparsity $\mathrm{S}$              & 0.2 \\
    minimum of mask changing $\eta_{min}$  & 0 \\
    maximum of mask changing $\eta_{max}$  & 100 \\
    balance factor   $\lambda$             & 10 \\
    mask changing interval   $t_m$        & 5000 \\
    threshold in Equation \ref{eq:unseen} & 25\\
    \hline
  \end{tabular}
\end{table}

\textbf{Training Resources}.
We use NVIDIA GeForce RTX 3090 to train each model.
The training duration for each model is typically observed to be 36 hours in the 50 tasks setting, while it takes approximately 24 hours in the 30 tasks setting.
However, since each environment needs to be trained three times with different seeds, the total training time is usually multiplied by three.

\section{ERK initialization}\label{app:ERK}
This section elucidates the utilization of the Erdős-Rényi Kernel (ERK), as proposed by \citet{ERK}, for initializing the sparsity in each layer of the model.
ERK tailors sparsity distinctively for different layers. 
In convolutional layers, the proportion of active parameters is determined by $\frac{n_{l-1}+n_l+w_l+h_l}{n_{l-1} \times n_l \times w_l \times h_l}$, where $n_{l-1}, n_l, w_l$, and $h_l$ represent the number of input channels, output channels, and the kernel's width and height in the $l$-th layer, respectively.
For linear layers, the active parameter ratio is set to $\frac{n_{l-1}+n_l}{n_{l-1} \times n_l}$, with $n_{l-1}$ and $n_l$ indicating the number of neurons in the $(l-1)$-th and $l$-th layers. 
ERK ensures that layers with fewer parameters maintain a higher proportion of active parameters.

\section{Baselines}\label{app:baseline}

We compare our proposed HarmoDT with the following baselines.
\begin{enumerate}[leftmargin=*, label=\textit{\roman*}.]
    \item \textbf{MTBC}. We extend Behavior cloning (BC) to multi-task offline policy learning via network scaling and a task-ID conditioned actor that is similar to MTIQL.
    \item \textbf{MTIQL}. We extend IQL \citep{IQL} with multi-head critic networks and a task-ID conditioned actor for multi-task policy learning. The TD-based baselines are used to demonstrate the effectiveness of conditional generative modeling for multi-task planning.
    \item \textbf{MTDIFF-P} \citep{he2023diffusion}. MTDIFF-P is a diffusion-based method that incorporates Transformer backbones and prompt learning for generative planning and data synthesis in multitask offline settings.
    \item \textbf{MTDT}. We extend the DT architecture \citep{DT} to learn from multitask data. Specifically, MTDT concatenates an embedding z and a state s as the input tokens, where z is the encoding of task ID. In evaluation, the reward-to-go and task ID are fed into the Transformer to provide task-specific information.
    \item \textbf{Prompt-DT} \citep{PDT}. Prompt-DT built on DT aims to learn from multi-task data and generalize the policy to unseen tasks. Prompt-DT generates actions based on the trajectory prompts and reward-to-go. 
\end{enumerate}

In addition to offline methods, our analysis also encompasses a comparison with several online methodologies to provide a comprehensive evaluation of our approach. These include:

\begin{enumerate}[leftmargin=*, label=\textit{\roman*}., start=6]
    \item \textbf{CARE} \citep{sodhani2021multi}. This method utilizes additional metadata alongside a combination of multiple encoders to enhance task representation, offering a nuanced approach to multi-task learning.
    \item \textbf{PaCO} \citep{sun2022paco}. PaCO introduces a parameter compositional strategy, ingeniously recombining task-specific parameters to foster a more flexible and adaptive learning process.
    \item \textbf{Soft-M} \citep{yang2020multi}.This approach is centered around the development of a routing network, which adeptly orchestrates the soft combination of various modules, thereby facilitating more dynamic learning pathways.
    \item \textbf{D2R} \citep{he2023not}. D2R innovatively employs disparate routing paths, enabling the selection of varying numbers of modules tailored to the specific requirements of each task, thereby enhancing the model's adaptability and efficiency.
\end{enumerate}

\begin{figure*}[ht!]
    \centering
    \subfigure{
    \centering
    \label{fig:s12}
    \includegraphics[width=0.23\textwidth]{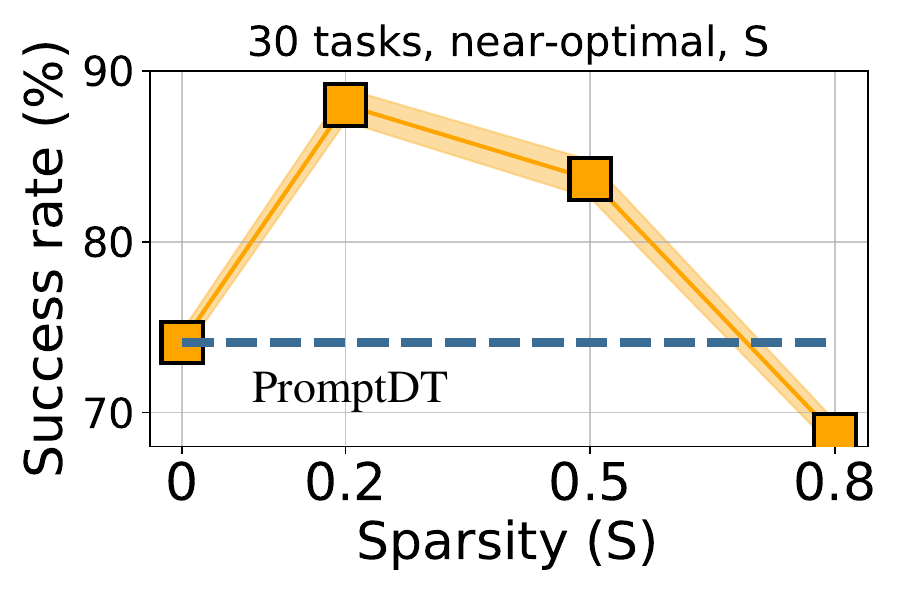}}
    %\hspace{2mm}
    \subfigure{
    \centering
    \label{fig:s22}
    \includegraphics[width=0.23\textwidth]{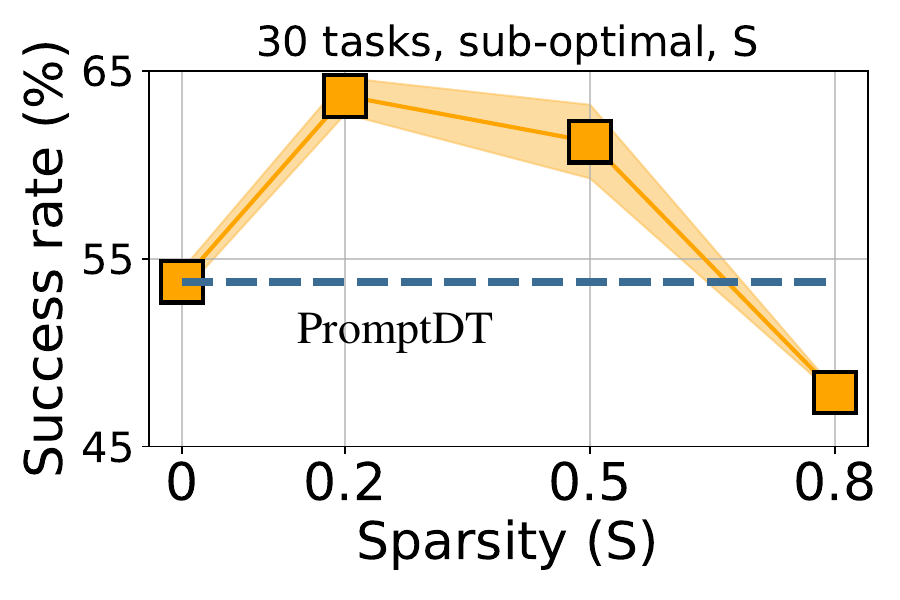}}
    \subfigure{
    \centering
    \label{fig:s32}
    \includegraphics[width=0.23\textwidth]{figs/50tasknear_sparsity.pdf}}
    \centering
    \subfigure{
    \centering
    \label{fig:s42}
    \includegraphics[width=0.23\textwidth]{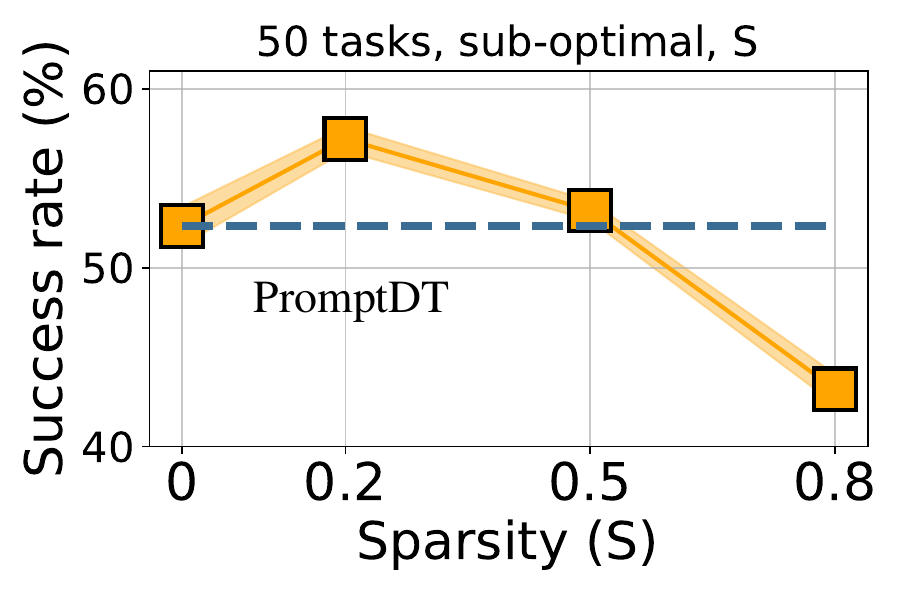}}
    \subfigure{
    \centering
    \label{fig:e12}
    \includegraphics[width=0.23\textwidth]{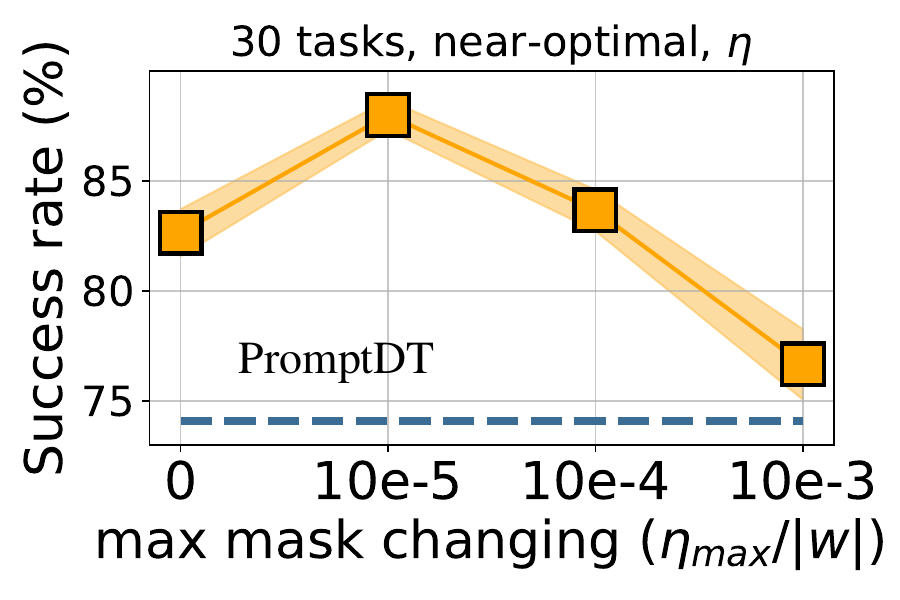}}
    \subfigure{
    \centering
    \label{fig:e22}
    \includegraphics[width=0.23\textwidth]{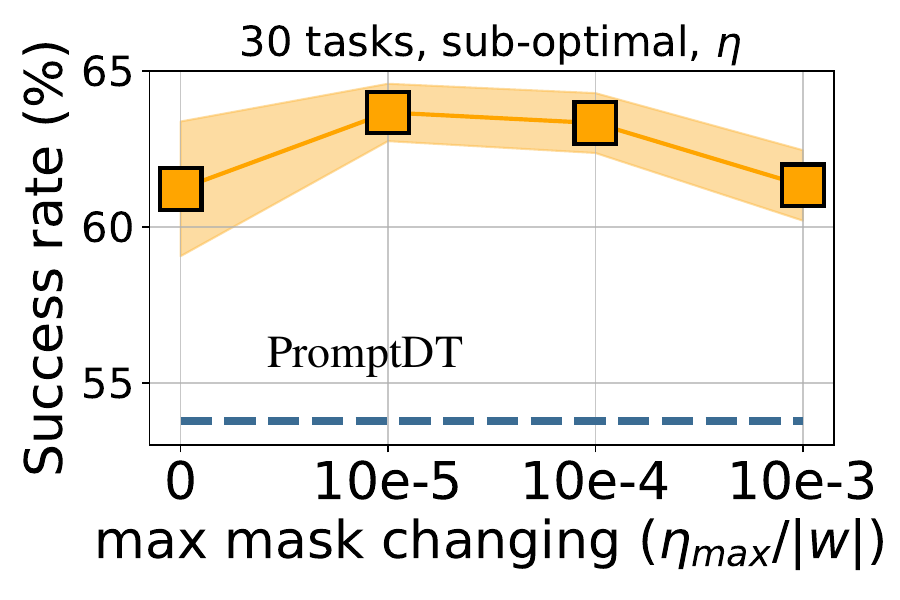}}
    \subfigure{
    \centering
    \label{fig:e32}
    \includegraphics[width=0.23\textwidth]{figs/50tasknear_eta.pdf}}
    \subfigure{
    \centering
    \label{fig:e42}
    \includegraphics[width=0.23\textwidth]{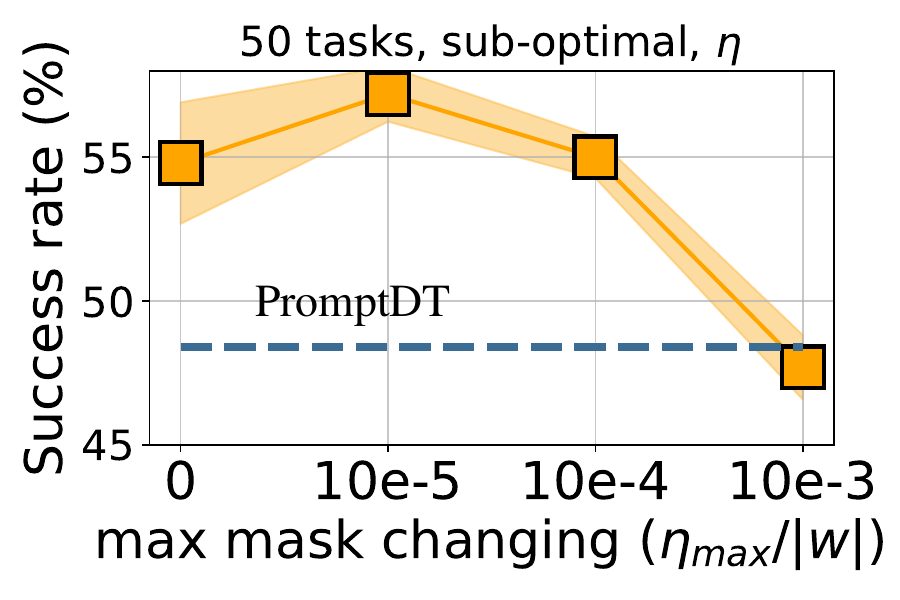}}
    % \vspace{-3cm}
    \subfigure{
    \centering
    \label{fig:g12}
    \includegraphics[width=0.23\textwidth]{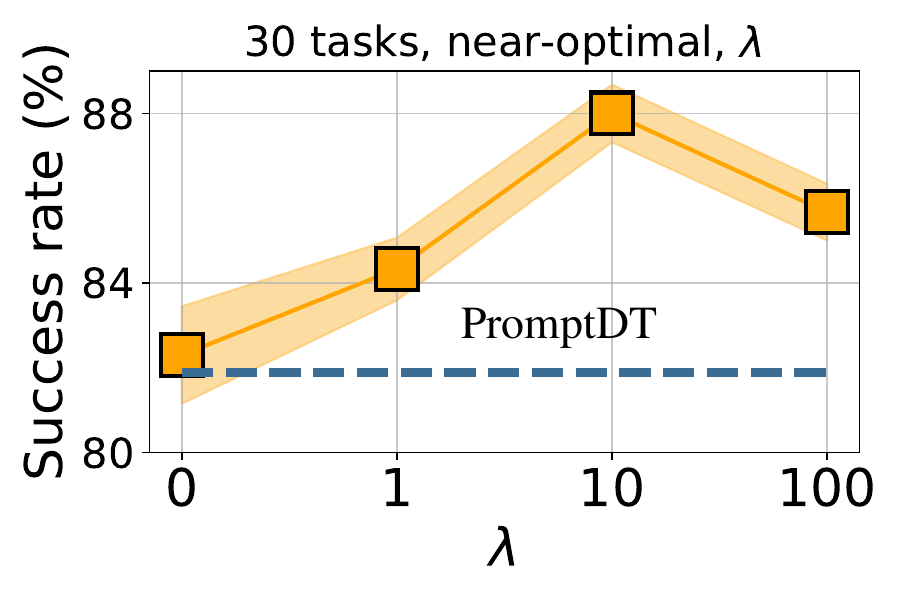}}
    \subfigure{
    \centering
    \label{fig:g22}
    \includegraphics[width=0.23\textwidth]{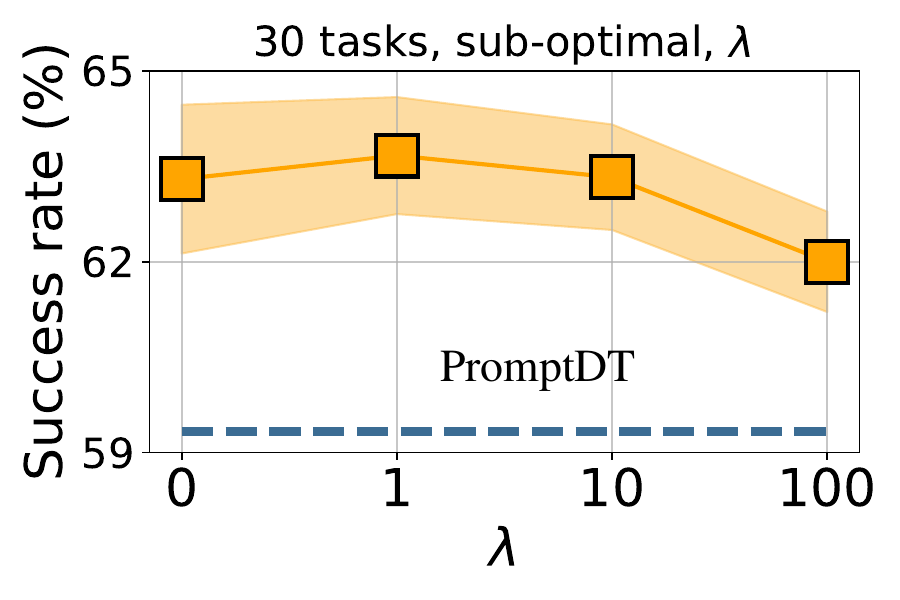}}
    \subfigure{
    \centering
    \label{fig:g32}
    \includegraphics[width=0.23\textwidth]{figs/50tasknear_lambda.pdf}}
    \subfigure{
    \centering
    \label{fig:g42}
    \includegraphics[width=0.23\textwidth]{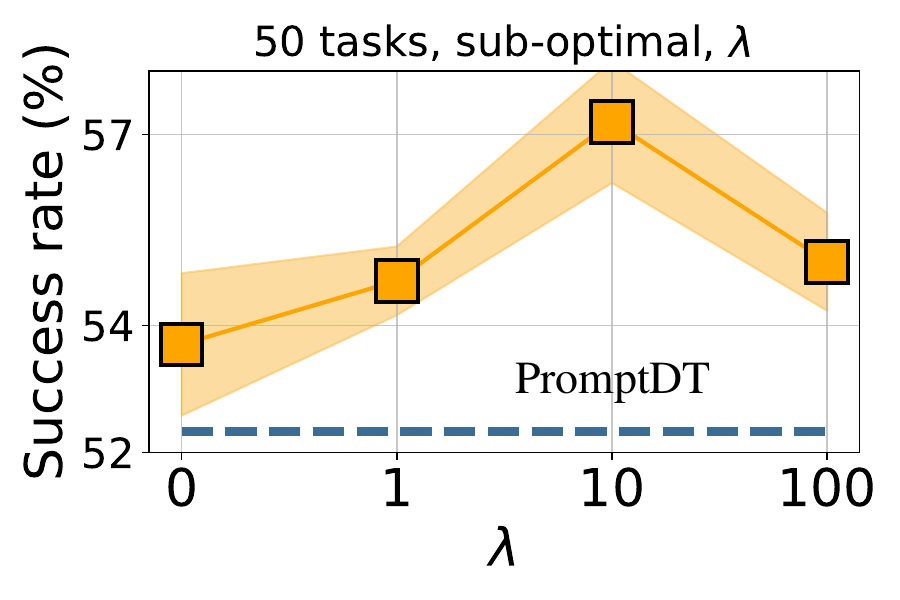}}
    %\hspace{2mm}
    \vspace{-.3cm}
    \caption{This figure presents an ablation study on critical hyper-parameters: sparsity ratio ($\mathrm{S}$), maximum mask change ($\eta_{max}$), and balance ratio ($\gamma$). Displayed from left to right are the results for 30 tasks (near-optimal and sub-optimal) and 50 tasks (near-optimal and sub-optimal). Default settings are $\eta_{max}=100$, $\mathrm{S}=0.2$, and $\lambda=10$. Each ablation varies one parameter while others remain default.}
    \label{fig:ablation_full}
    \vspace{-.3cm}
\end{figure*}

\section{More Ablations}\label{sec:moreab}
This section comprehensively details the ablation study conducted on key hyper-parameters within our experimental framework. These hyper-parameters include $\eta_{max}$, which is integral to the cosine annealing process (with a fixed $\eta_{min}=0$), the mask alteration frequency denoted as $t_m$, the overall sparsity parameter $\mathrm{S}$, and the balance controller $\lambda$.
Figure~\ref{fig:ablation_full} presents an in-depth analysis of these hyper-parameters' impact on the performance of our model across 30 and 50 tasks within the Meta-world benchmark. This evaluation spans both near-optimal and sub-optimal settings, providing a comprehensive understanding of the hyper-parameters' influence under varied conditions. 
Remarkably, our approach consistently surpasses baseline models across a diverse range of hyper-parameter values. From this extensive analysis, we derive optimal settings for these parameters: a sparsity ratio $\mathrm{S}$ set to 0.2, an $\eta_{max}=100$ value approximating 0.001\% of the total weight count, a balance factor $\lambda$ fixed at 10, and a mask changing interval $t_m$ established at 5000 rounds. These recommended configurations are grounded in empirical evidence and are instrumental in achieving enhanced performance in multi-task learning scenarios.

\section{Related Work}

\subsection{Offline Reinforcement Learning}

Offline RL algorithms learn a policy entirely from this static offline dataset $\gD$, without online interactions with environment \citep{levine2020offline}. 
This paradigm can be precious in case the interaction with the environment is expensive or high-risk (e.g., safety-critical applications).
However as the learned policy might differ from the behavior policy, the offline algorithms must mitigate the effect of the \textit{distribution shift}, which can result in a significant performance drop, as demonstrated in prior research \citep{fujimoto2019off}.
Several previous works have utilized constrained or regularized dynamic programming to mitigate deviations from the behavior policy \citep{TD3BC, CQL, IQL}.

Conditional sequence modeling approaches have been a promising direction for solving offline RL, which predicts subsequent actions from a sequence of past experiences, encompassing state-action-reward triplets.
This paradigm lends itself to a supervised learning approach, inherently constraining the learned policy within the boundaries of the behavior policy and focusing on a policy conditioned on specific metrics for future trajectories \citep{DT, GDT, QT, QDT, PTDT, CommFormer, meng2023offline}.

Recently, there has been a growing interest in incorporating diffusion models into offline RL methods. 
This alternative approach to decision-making stems from the success of generative modeling, which offers the potential to address offline RL problems more effectively.
Diffuser and its variants \citep{Diffuser, DD, chen2022offline, DQL}
utilize diffusion-based generative models to represent policies or model dynamics, achieving competitive or superior performance across various tasks.

\subsection{Multi-Task Reinforcement Learning}

Multi-task RL aims to learn a shared policy for a diverse set of tasks, and there are many different approaches have been proposed in the literature \citep{xu2020knowledge, yang2020multi, sarafian2021recomposing, sodhani2021multi}.
One of the most straightforward approaches to MTRL is to formulate the multi-task model as a task-conditional one \citep{yu2020meta}, as commonly used in goal-conditional RL \citep{plappert2018multi}.
Conditional sequence modeling approaches, which consider handling multi-task problems, mainly rely on expert trajectories and entail substantial training expenses \citep{PDT, PTDT, lee2022multi}.
Diffusion model is also verified to have the potential to address the challenge of multi-task generalization in RL.
MTDIFF \citep{he2023diffusion} extends the conditional diffusion model to be capable of solving multi-task decision-making problems and synthesizing useful data for downstream tasks.

Although these methods are simple and have shown some success in certain cases, one inherent limitation is the conflicting gradients among different data sources,  phenomenon known as gradient conflicts and
negative transfer in many fields, such as federated learning~\citep{fed1,fed2,fed3,fed4}, domain generalization~\citep{dg1,dg2,dg3}, and multi-task learning~\citep{yu2020gradient,chen2020just,liu2021conflict}.
To mitigate conflicting gradient impacts in a multi-task context, several methodologies have been developed: PCGrad \citep{yu2020gradient} projects each task's gradient onto the orthogonal plane of another's, subsequently updating parameters using the mean of these projected gradients. Graddrop \citep{chen2020just} employs a stochastic approach, randomly omitting certain gradient elements based on their conflict intensity. CAGrad \citep{liu2021conflict} manipulates gradients to converge towards a minimum average loss across tasks.
In contrast, our method, HarmoDT, leverages gradient information in a fundamentally different manner. Instead of adjusting gradients post hoc as in these methods, we proactively utilize gradient data to inform the selective activation of parameters for each task through a masking mechanism. This direct intervention at the parameter level allows the model to update without the typical interferences found in gradient-level adjustments, fostering a more streamlined and potentially more efficacious optimization process.

On the other hand, \citet{d2020sharing} leverages the shared knowledge between multiple tasks by using a shared network followed by multiple task-specific heads.
\citet{yang2020multi} further extends this approach by softly sharing features (activations) from a base network among tasks, by generating the combination weight with an additional modularization network taking both state and task-id as input.
Since the base and modularization networks take state and task information as input, there is no clear separation between task-agnostic and task-specific parts.
PaCo \citep{sun2022paco} explores a compositional structure in the parameter space and distinguishes the task-agnostic and task-specific parts with different parameters, however, it still suffers the conflicting gradients within the shared parameters.
Our method uses task-specific masks to find out the task-agnostic and task-specific parameters and dynamically update them to mitigate the conflicting gradients phenomenon, achieving state-of-the-art performance across various benchmarks.

%%%%%%%%%%%%%%%%%%%%%%%%%%%%%%%%%%%%%%%%%%%%%%%%%%%%%%%%%%%%%%%%%%%%%%%%%%%%%%%
%%%%%%%%%%%%%%%%%%%%%%%%%%%%%%%%%%%%%%%%%%%%%%%%%%%%%%%%%%%%%%%%%%%%%%%%%%%%%%%

\end{document}